\documentclass{article}
\usepackage{arxiv}

\title{Statistical quantification of confounding bias in predictive modelling}

\author{
  Tamas~Spisak \\
  Institute for Diagnostic and Interventional Radiology and Neuroradiology \\
  University Hospital Essen\\
  Hufelandstrasse 55, 45147 Essen \\
  \texttt{tamas.spisak@uk-essen.de} \\
}

\begin{document}
\maketitle

\begin{abstract} 
The lack of non-parametric statistical tests for confounding bias significantly hampers the development of robust, valid and generalizable predictive models in many fields of research.

Here I propose the \emph{partial} and \emph{full confounder tests}, which, for a given confounder variable, probe the null hypotheses of \emph{unconfounded} and \emph{fully confounded models}, respectively.

The tests provide a strict control for Type I errors and high statistical power, even for non-normally and non-linearly dependent predictions, often seen in machine learning.
Applying the proposed tests on models trained on functional brain connectivity data from the Human Connectome Project and the Autism Brain Imaging Data Exchange dataset reveals confounders that were previously unreported or found to be hard to correct for with state-of-the-art confound mitigation approaches.

The tests (implemented in the package \emph{mlconfound}\footnote{\href{https://mlconfound.readthedocs.io}{https://mlconfound.readthedocs.io}}) can aid the assessment and improvement of the generalizability and neurobiological validity of predictive models and, thereby, foster the development of clinically useful machine learning biomarkers.
\end{abstract}

\keywords{machine learning, predictive modelling, confounding bias, confounder test, conditional independence, conditional permutation}

\section{Introduction}

Predictive modelling has recently become increasingly important in biomedical research and holds promise for delivering biomarkers that substantially impact clinical practice and public health\citep{vogt2018machine, kent2018personalized, spisak2020pain, walsh2021dome}. When evaluating the usefulness and applicability of such markers, predictive performance is far from being the only important consideration. Biomedical validity (i.e. whether the model is driven by biomedically relevant signal) as well as generalizability and fairness across contexts and populations are also fundamental requirements for candidate biomarkers\citep{woo2017building, obermeyer2019dissecting, mehrabi2021survey}.

Spurious, out-of-interest associations between the predictor variables (features) and the prediction target can be detrimental to the model's biomedical validity and generalizability. This phenomenon is often called confounding bias\citep{prosperi2020causal}. Confounding bias can be driven by - among others - measurement artifacts (e.g motion artifacts in magnetic resonance imaging-based predictive models of Alzheimer's\citep{rao2017predictive}, attention deficit hyperactivity disorder\citep{eloyan2012automated, couvy2016head} or Autism Spectrum Disorder (ASD)\citep{gotts2013perils, spisak2014voxel, spisak2019optimal}), demographic and psychometric variables (e.g models trained to predict intelligence\citep{ cole2012global, he2020deep} might provide a statistically significant predictive performance by picking up solely on age-related variance\citep{dubois2018distributed, lohmann2021predicting}), sampling bias and stochastic group differences (e.g. racially biased machine learning models\citep{ obermeyer2019dissecting, lwowski2021risk}) as well as batch effects or, in multi-center studies, center-effects.

While various data cleaning methods may help in mitigating confounding bias\citep{rao2017predictive, dukart2011age, spisak2014voxel, abdulkadir2014reduction, johnson2007adjusting}, it is often unclear which variables should be considered as confounders and such approaches hold risks of eliminating signal-of-interest\citep{wachinger2021detect}.

\begin{figure}[!b]
  \centering
  \includegraphics[width=0.5\paperwidth]{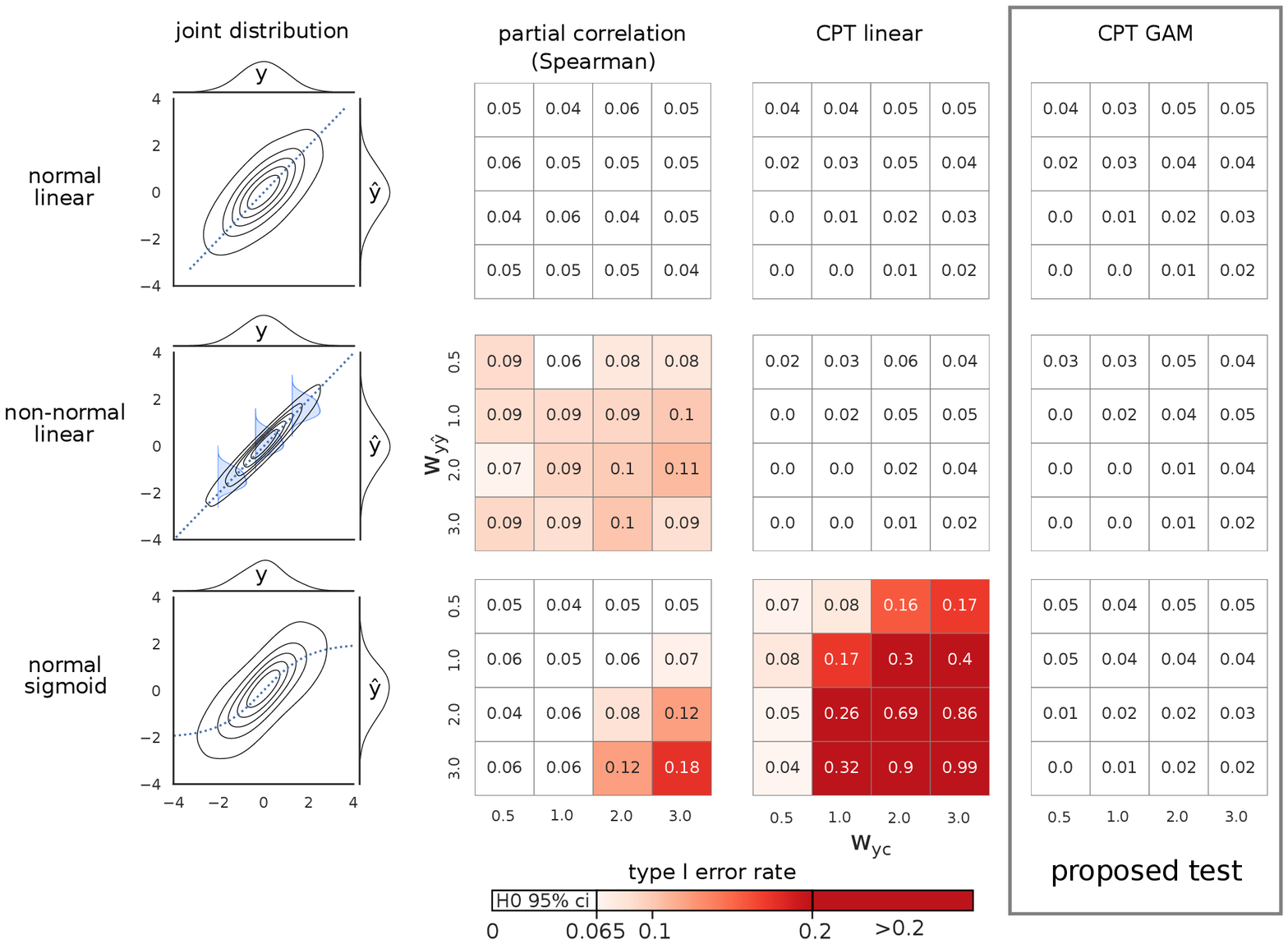}
  \caption{\textbf{Type I error control of partial Spearman correlation, linear and GAM-based conditional permutation test}. \\
  Type I error control was investigated in three example cases: normal conditional distribution with linear dependency (first row), slightly non-normal conditional distribution with linear dependency (second row) and normal conditional distribution with non-normal (sigmoid) dependency (third row). Non-normal conditional distribution on the second plot is illustrated with blue density diagrams (kurtosis: -0.8, skewness: -0.1). False positive rates for confounder contributions ($w_{yc}$) and predictive performances ($w_{y\hat{y}}$) is shown in heatmaps. The upper limit for the binomial confidence interval corresponding to $alpha=0.05$ is 0.065. Values below this threshold (colored white) indicate a valid type I error control.
  }
  \label{fig:sim-h0-demo}
\end{figure}

Powerful and robust statistical tests for quantifying confounding bias in predictive models could substantially foster both the identification of confounders to correct for and the assessment of the effectiveness of various confound-mitigation approaches. It is tempting to think about confounding bias as the \emph{conditional dependence} of the model predictions on the confounder, given the target variable. However, the proper evaluation of conditional independence among these variables is challenging. Namely, even in the presence of a slight non-normality and/or non-linearity of the involved conditional distributions, the 'conditional' analogs of the most popular bivariate non-parametric tests (like the partial Spearman correlation, see Fig. \ref{fig:sim-h0-demo}) are not valid measures of conditional independence. Although warnings about this issue were given from early on\citep{korn1984ranges}, and received a fair amount of attention recently\citep{bergsma2010nonparametric, candes2016panning, peters2016causal,  shah2020hardness, berrett2020conditional}, the magnitude of the problem may not be fully appreciated in case of predictive model diagnostics, where non-normality and non-linearity of the model output can be frequently seen, as a consequence of e.g. feature-set characteristics and model regularization\citep{garcia2009study, kristensen2017whole} (see Supplementary Material \ref{sup:nomlinviol} for a simplistic example).

Recently, two different approaches were proposed for quantifying confounding bias \citep{chaibub2019permutation, ferrari2020measuring}. However, these methods either fail to control type I error (as known in the case of balanced permutations\cite{southworth2009properties, hemerik2018exact}, used in ref.\cite{chaibub2019permutation}), or do not provide p-values at all\cite{ferrari2020measuring}. 
Moreover, without some modifications, they are only applicable for categorical variables and involve re-fitting the model, which may not be feasible for models with high computational cost (e.g. when trained with nested cross-validation).

This work aims to construct a statistical test for confounding bias that (i) guarantees valid type-I error control for arbitrary models, even if non-normal and/or non-linear dependencies are involved (ii) does not require re-fitting the model, (iii) is applicable for classification as well as for prediction problems and both with numerical and categorical confounders.

\section{Results}

\begin{figure}[!b]
  \centering
  \includegraphics[width=0.6\paperwidth]{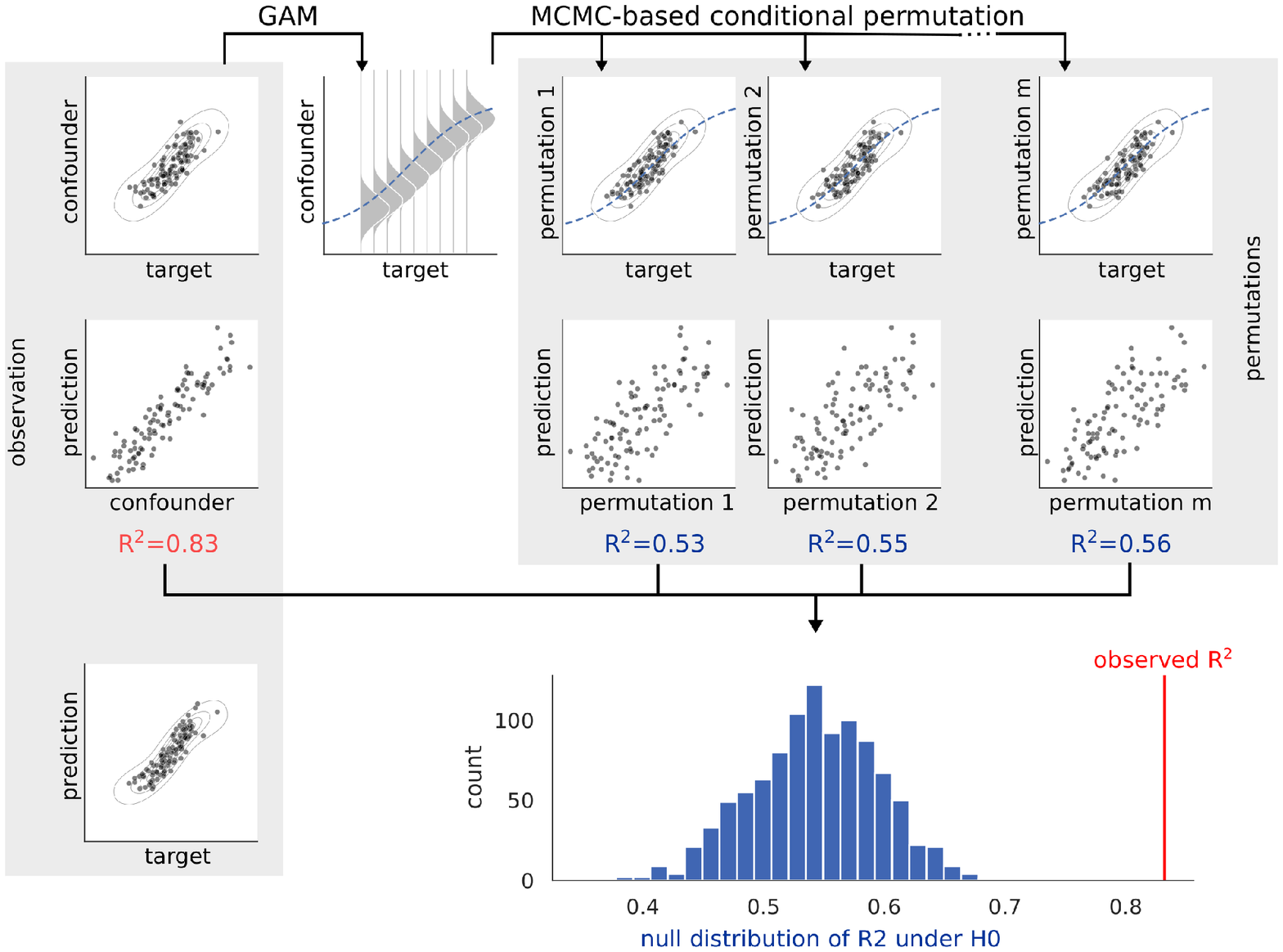}
  \caption{\textbf{Graphical representation of the proposed partial confounder test}. \\
    The partial confounder test models the conditional distribution of the confounder, given the target variable, with a generalized additive model (GAM). The parallel-pairwise Markov-chain Monte-Carlo (MCMC) sampler draws permutations of the confounder variable that comply with the GAM-based conditional distribution (permutation 1, 2, ..., m). The test statistic (coefficient of determination, $R^2$) is then computed between the model predictions and the original, as well as the permuted confounder variables. The original and the permuted test statistics construct the p-value as the ratio of permuted test statistics more extreme than the original. Figure source code available as jupyter notebook: \href{https://github.com/pni-lab/mlconfound-manuscript/blob/main/simulated/overview-fig.ipynb}{https://github.com/pni-lab/mlconfound-manuscript/blob/main/simulated/overview-fig.ipynb}
  }
  \label{fig:overview}
\end{figure}


\subsection{The partial and full confounder tests}

The concept of conditional independence provides a straightforward framework for assessing confounding bias in predictive models. However, handling the non-normal and/or non-linear conditional dependencies often seen in predictive models\citep{garcia2009study, kristensen2017whole} poses a great challenge.
In fact, as recently shown by Shah and colleagues in their 'no free lunch' theorem\cite{shah2020hardness}, it is effectively impossible to establish a \emph{fully non-parametric} conditional independence test with a valid type I error control and a non-trivial power. Indeed, perhaps somewhat surprisingly, but not totally unexpectedly\cite{korn1984ranges} - partial correlation-like analogs of widely used bivariate non-parametric test, like partial Spearman correlation, exhibit inflated type I errors even with slight violations of normality and/or linearity (as clearly demonstrated with simulated data on Fig. \ref{fig:sim-h0-demo}). Such tests are therefore poor choices for testing confounding bias in machine learning.

As, in terms of its conditional distribution on the others, the model output is clearly the most intractable from the three involved variables (target, prediction, confounder)\citep{garcia2009study, kristensen2017whole}, a method being distribution-free only for this variable may already provide a sufficient robustness for predictive model diagnostics. Exactly this can be achieved with the proposed approach, which extends the novel framework of conditional permutation testing (CPT)\cite{berrett2020conditional} with conditional distribution estimation via generalized additive (GAM)\citep{hastie1987generalized} or multinomial logistic models (mnlogit)\cite{bennett1966multiple, jones1975proability}. The proposed approach offers two novel tests for probing confounding bias: the \emph{full confounder test} probes whether the model's predictive performance can be exclusively attributed to the confounder and the \emph{partial confounder test} investigates whether the model utilizes any confounder-variance in the predictions, when controlled for the target variable.
These tests place no assumptions on the conditional distributions of the model predictions, ensuring valid model diagnostics even in cases of non-normally and non-linearly dependent predictions.

The inner workings of the \emph{partial confounder test} are summarized on Fig. \ref{fig:overview}. In short, the test models the conditional distribution between the confounder and the target variable with a GAM - or with an \emph{mnlogit} regression, in case of categorical confounder - and then uses a so-called parallel-pairwise Markov-chain Monte-Carlo sampler\cite{berrett2020conditional} that draws permutations of the confounder, so that the permuted variables still comply with the estimated conditional distribution. The test statistic (coefficient of determination, $R^2$) is then computed between the model predictions and the original, as well as the permuted variables. The original and the permuted test statistics construct the p-value as the ratio of permuted test statistics more extreme than the original. The \emph{full confounder test} works in an analogous way, with the difference that it creates permuted copies of the target variable, instead of the confounder.

These tests can handle both numerical and categorical data and can be applied for any predictive model, without needing to re-fit the model, that is, with a negligible extra computational cost.
The proposed \emph{partial} and \emph{full confounder tests} have been implemented in the python package \emph{mlconfound}\footnote{\href{https://mlconfound.readthedocs.io}{https://mlconfound.readthedocs.io}}. 

\begin{figure}[!b]
  \centering
  \includegraphics[width=0.6\paperwidth]{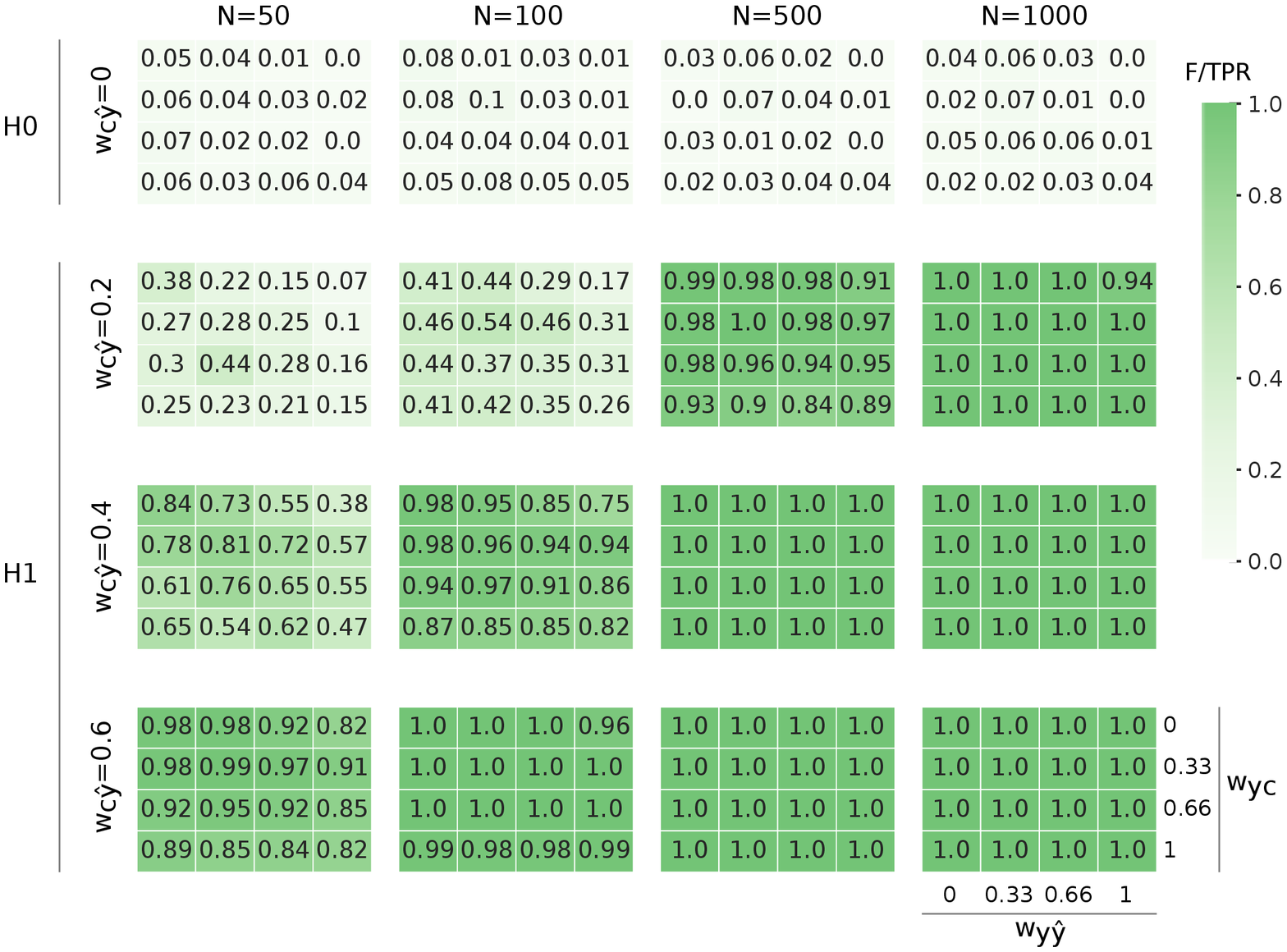}
  \caption{\textbf{Type I error control and power of the partial confounder test based on simulations with normal conditional distribution and linear dependencies.} \\
  Heatmaps depict positive rates (ratio of p-values lower than 0.05, color coded as shown by the palette on the right) in various simulations settings (100 simulations per tile) with different simulation weights $w_{y\hat{y}}$ (predictive performance; horizontal axis on each heatmap), $w_{yc}$ (confounder-target association; vertical axis on each heatmap), $w_{c\hat{y}}$ (degree of confoudner bias; rows) and for different sample sizes (N, columns). Weights 0.2, 0.33, 0.4, 0.6, 0.66, 1.0 can be assigned to the following approximate explained variance values: 4\%, 10\%, 12\%, 25\%, 30\%, 50\%, respectively. First row contains simulations under the null hypothesis (H0, no confounder bias), rows 2-4 represent simulations from the alternative hypothesis (H1, confounding bias).
  Positive rates for the simulations under the null and the alternative hypotheses can be interpreted as type I error rate and statistical power, respectively. The higher 95\% confidence limit for a positive rate of $alpha=0.05$ is 0.11 for each tile.
  }
  \label{fig:sim-normal}
\end{figure}

\subsection{Simulations}

\begin{figure}[!b]
  \centering

  \includegraphics[width=0.5\paperwidth]{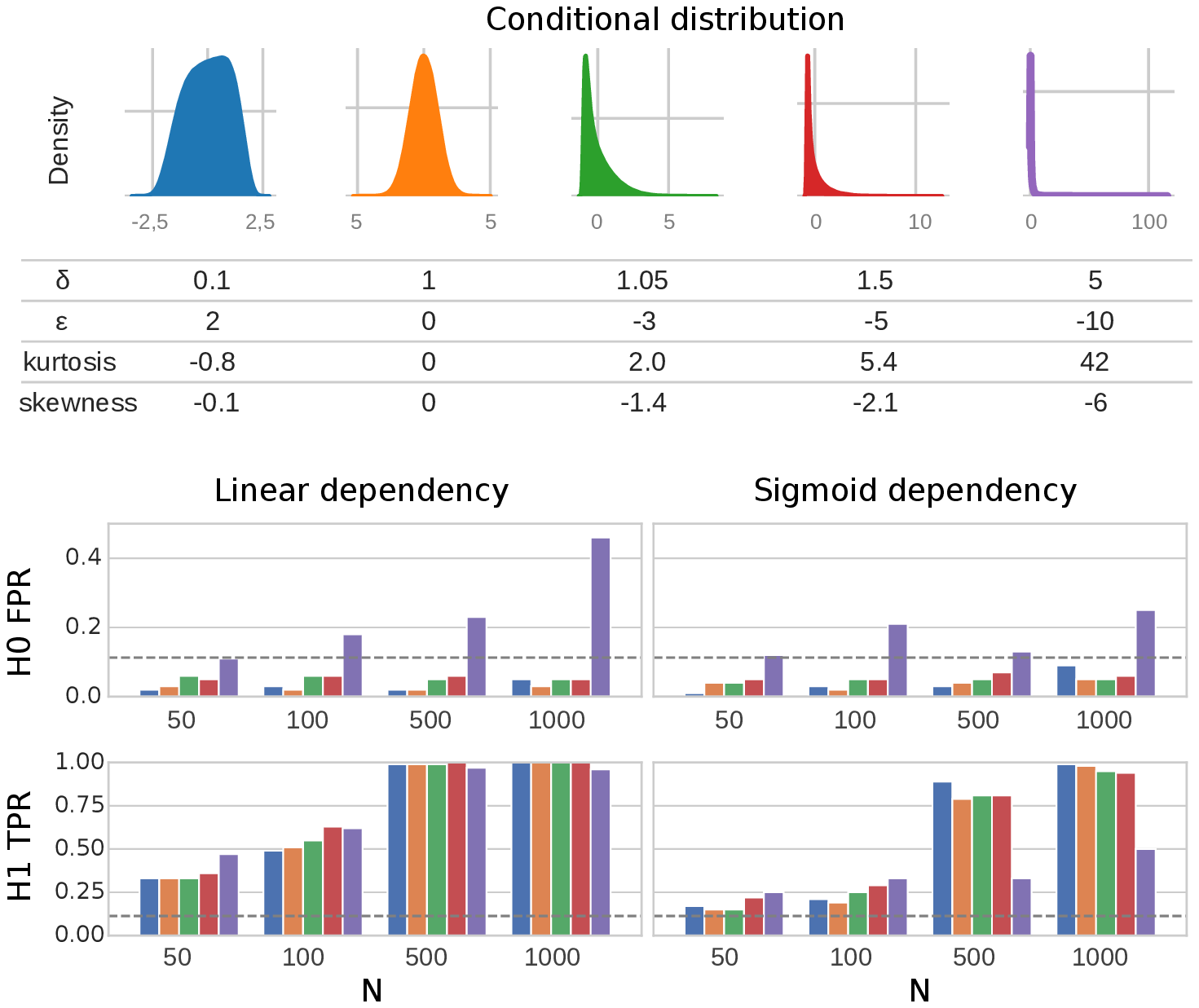}
  \caption{\textbf{Robustness of conditional permutation based confound testing to non-normality and non-linearity.} \\
  Simulations included variables with five different degrees of non-normality (top panel), as introduced with various $\delta$ and $\epsilon$ values of the \emph{sinh-arcsinh} transformation (yellow: normally distributed). Fisher's kurtosis and skewness is given for each distribution. False and true positive rates in the simulations under H0 and H1, respectively, for each investigated sample size (N), are depicted by barplots for both linear and sigmoid dependency structure. Upper 95\% binomial confidence limit corresponding to $alpha=0.05$ is shown with a vertical dashed line.}
  \label{fig:sim-non-normal}
\end{figure}

\subsubsection*{Type I error}

\begin{figure}[H]
  \centering
  \fbox{
   \includegraphics[width=0.4\paperwidth]{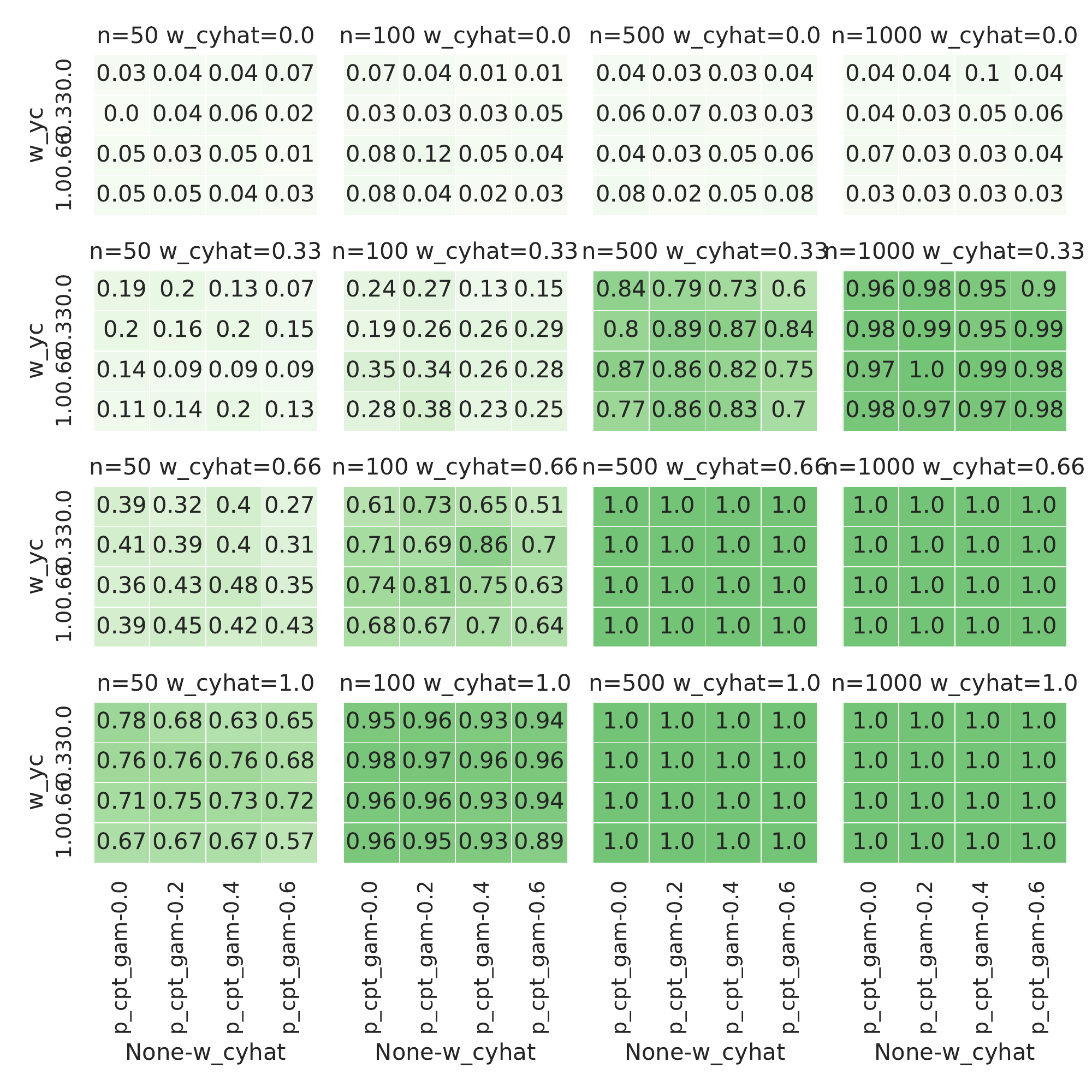}}
  \caption{Heatmaps showing the positive rates of the 'full' confounder test, with numerical variables, normal conditional distribution and linear dependence.}
  \label{fig:sim-ccc-lin-full}
\end{figure}

\begin{figure}[H]
  \centering
  \fbox{
   \includegraphics[width=0.4\paperwidth]{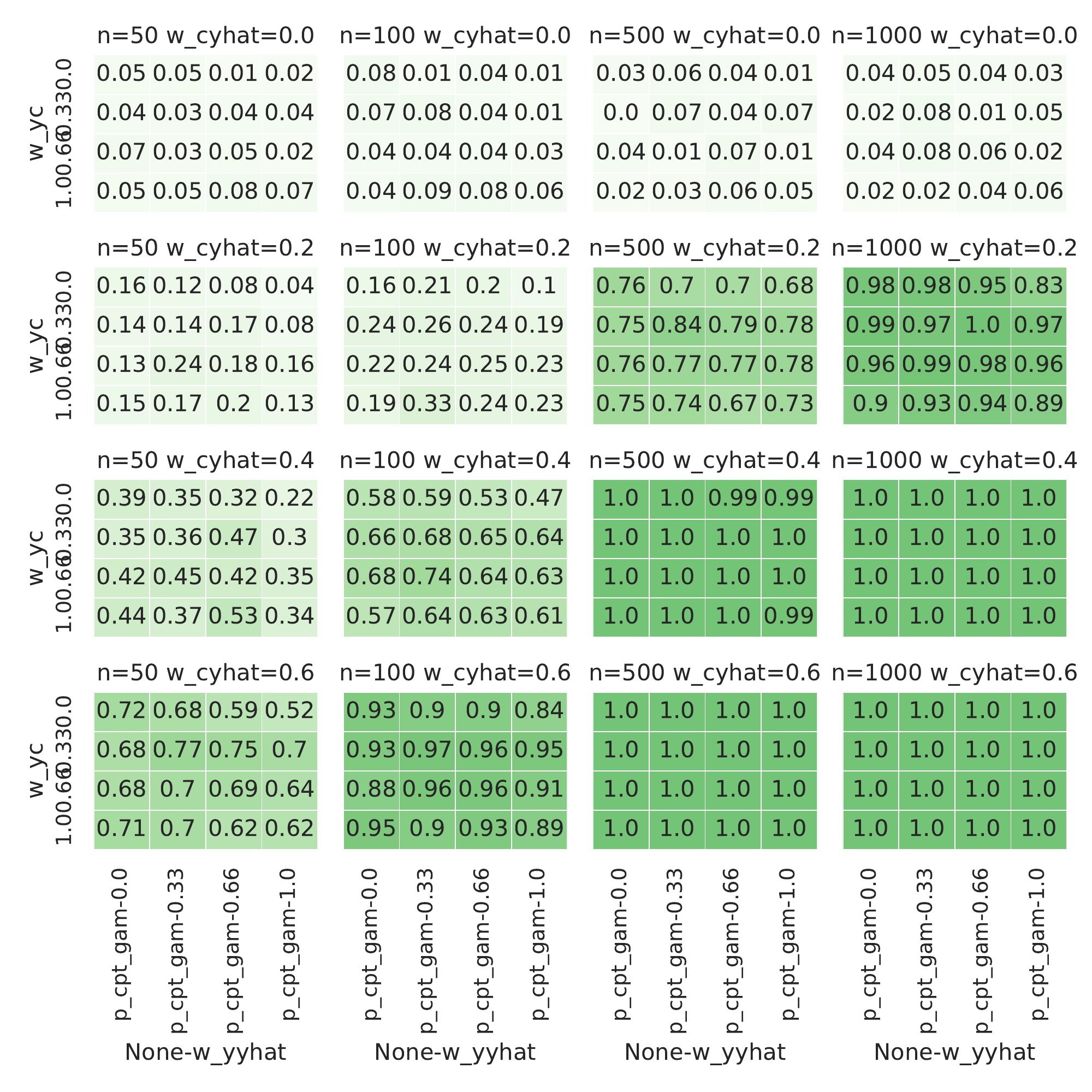}}
  \caption{Heatmaps showing the positive rates of the 'partial' confounder test, with numerical variables, normal conditional distribution and sigmoid dependence.}
  \label{fig:sim-ccc-sig-partial}
\end{figure}

\begin{figure}[H]
  \centering
  \fbox{
   \includegraphics[width=0.4\paperwidth]{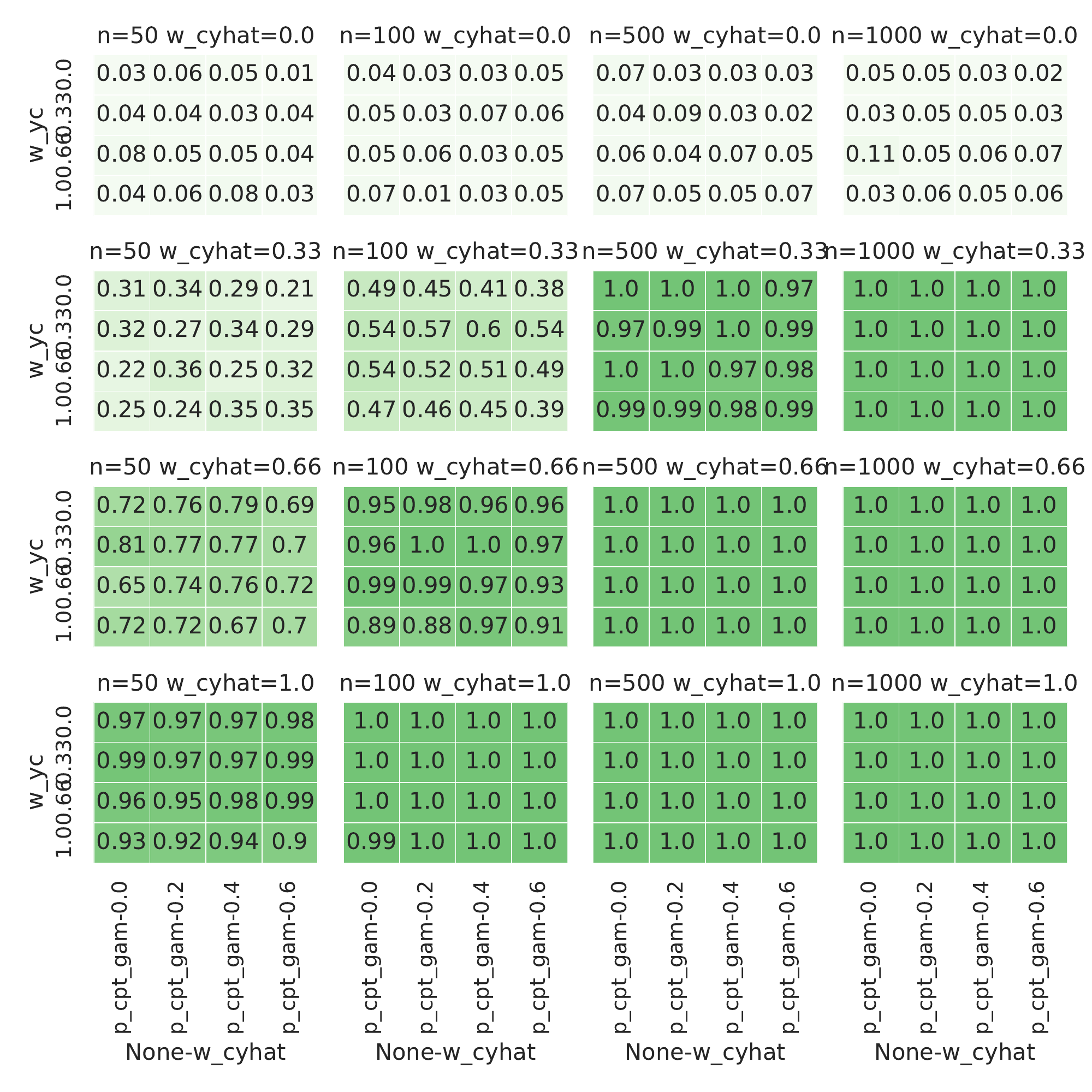}}
  \caption{Heatmaps showing the positive rates of the 'full' confounder test, with numerical variables, normal conditional distribution and sigmoid dependence.}
  \label{fig:sim-ccc-sig-full}
\end{figure}

As suggested by theory (see \hyperref[sec:methods]{Methods} \ref{sec:methods-pct} for details) and shown by the simulations with a wide range of settings, both of the proposed tests provide a valid Type I error control (Fig. \ref{fig:sim-normal} and Supplementary Figures \ref{fig:sim-bbb-lin-partial}, \ref{fig:sim-ccc-lin-full} and \ref{fig:sim-bbb-lin-full}), even in case of non-linearity and non-normality (Figs. \ref{fig:sim-h0-demo}, \ref{fig:sim-non-normal} and Supplementary Figures \ref{fig:sim-ccc-sig-partial}, \ref{fig:sim-bbb-sig-partial}, \ref{fig:sim-ccc-sig-full}, \ref{fig:sim-bbb-sig-full}), except when non-normality is very extreme (purple distribution on Fig. \ref{fig:sim-non-normal}, kurtosis: 42, skewness: -6).

\subsubsection*{Power}

Estimates of statistical power for the partial confounder test (with normal and linear simulations, for  a wide range of parameters) are shown on Figure \ref{fig:sim-normal}. Notably, with sample sizes as large as 1000, a confounder contributing only $\sim 4\%$ to the variance of the predictions  ($w_{c\hat{y}} = 0.2$) can already be robustly detected with a power of 94-100\%. With a sample size of 500, the same confounding bias is still detected with a power greater than 84-100\% in all of the simulation cases. A sample size of 100 requires a somewhat stronger bias with approximately 12\% of explained variance ($w_{c\hat{y}}=0.4$) to achieve a reasonable level of power (75-98\%). Finally, even with a relatively low sample size of 50, the same amount of confounder variance is detected with a power of at least 50\%. If the confounder explains more than 25\% of variance, it is almost certainly detected even with a low sample size of $n \geq 50$.

\begin{figure}[H]
  \centering
  \fbox{
   \includegraphics[width=0.4\paperwidth]{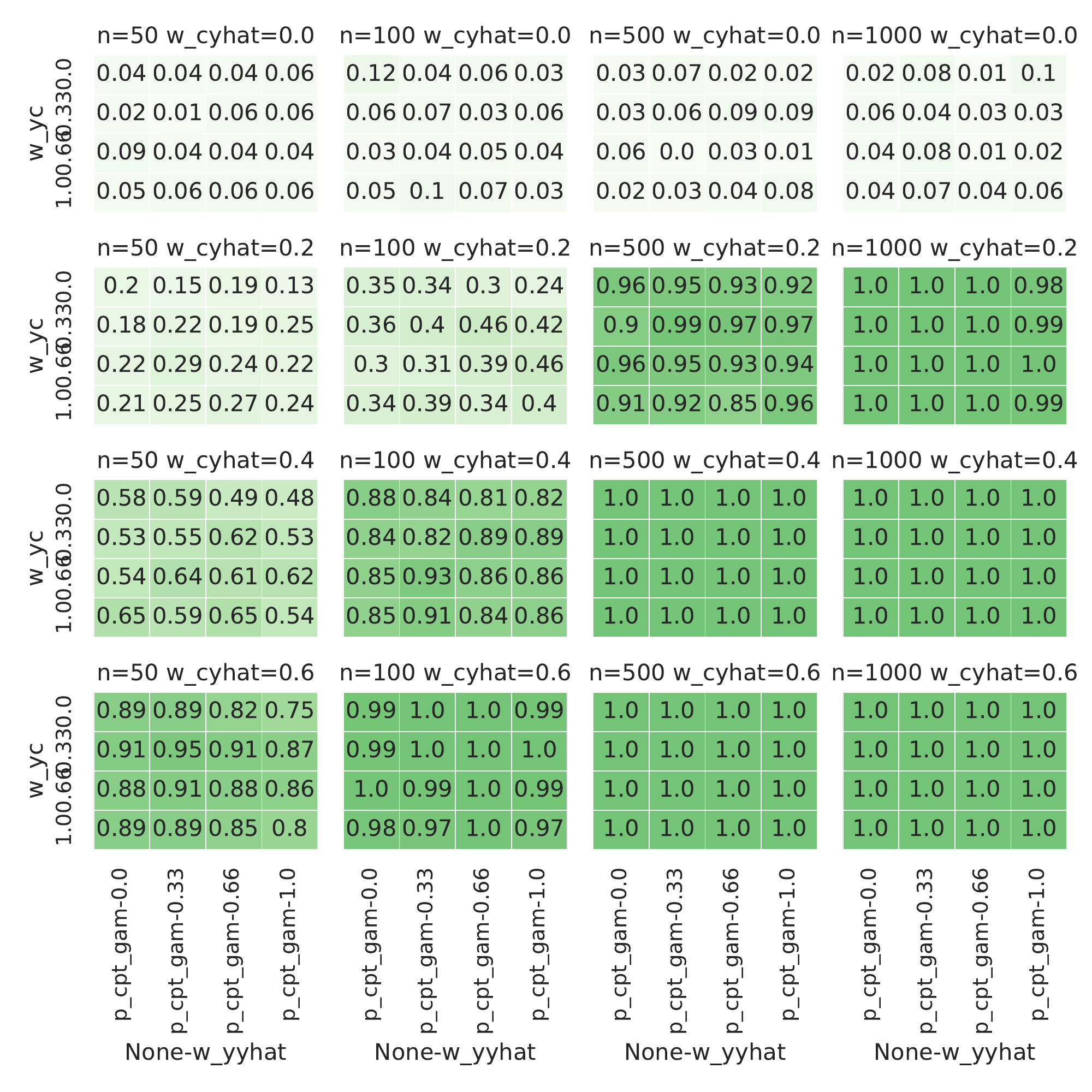}}
  \caption{Heatmaps showing the positive rates of the 'partial' confounder test, with categorical variables, normal conditional distribution and linear dependence.}
  \label{fig:sim-bbb-lin-partial}
\end{figure}

\begin{figure}[H]
  \centering
  \fbox{
   \includegraphics[width=0.4\paperwidth]{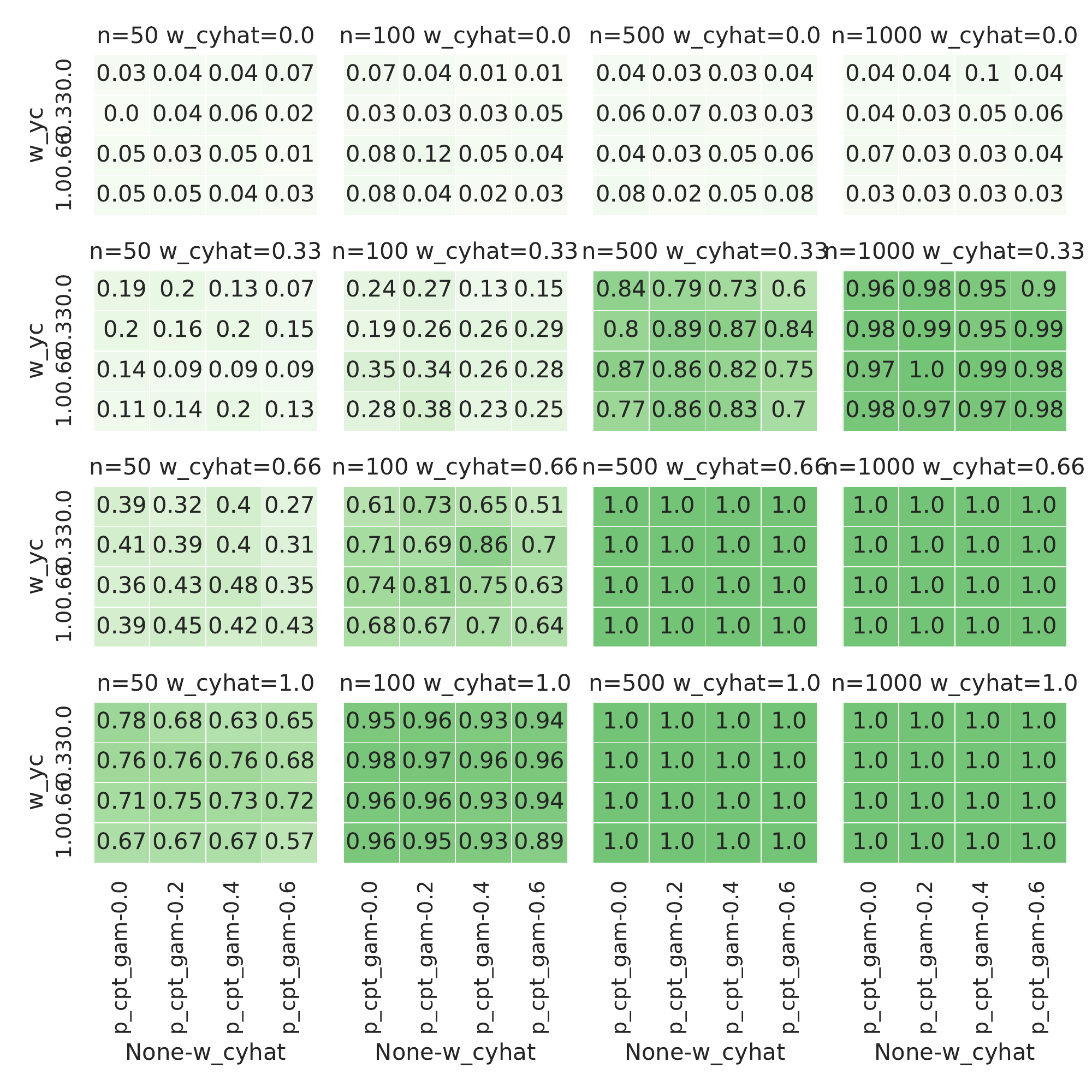}}
  \caption{Heatmaps showing the positive rates of the 'full' confounder test, with categorical variables, normal conditional distribution and linear dependence.}
  \label{fig:sim-bbb-lin-full}
\end{figure}

\begin{figure}[H]
  \centering
  \fbox{
   \includegraphics[width=0.4\paperwidth]{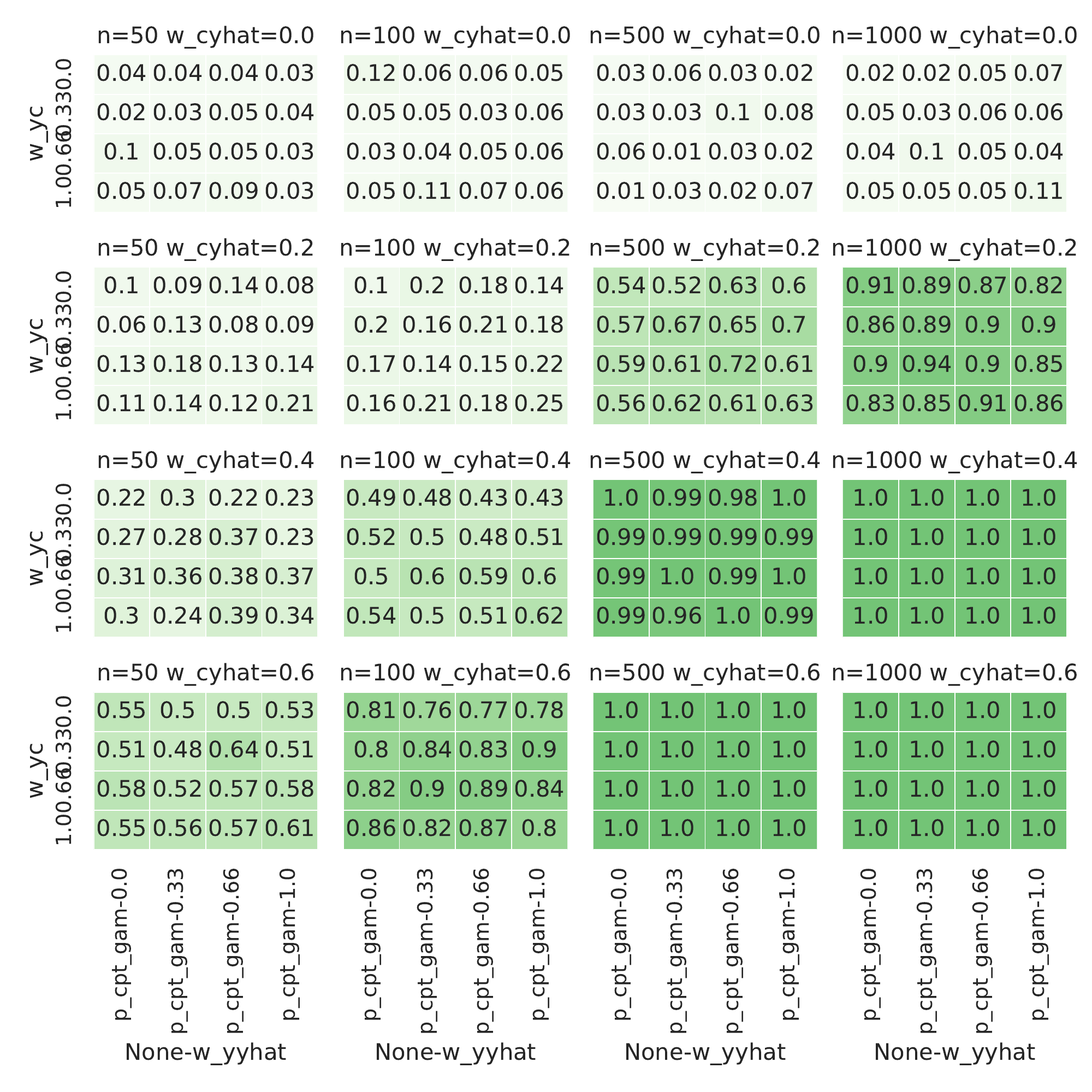}}
  \caption{Heatmaps showing the positive rates of the 'partial' confounder test, with categorical variables, normal conditional distribution and sigmoid dependence.}
  \label{fig:sim-bbb-sig-partial}
\end{figure}

\begin{figure}[H]
  \centering
  \fbox{
   \includegraphics[width=0.4\paperwidth]{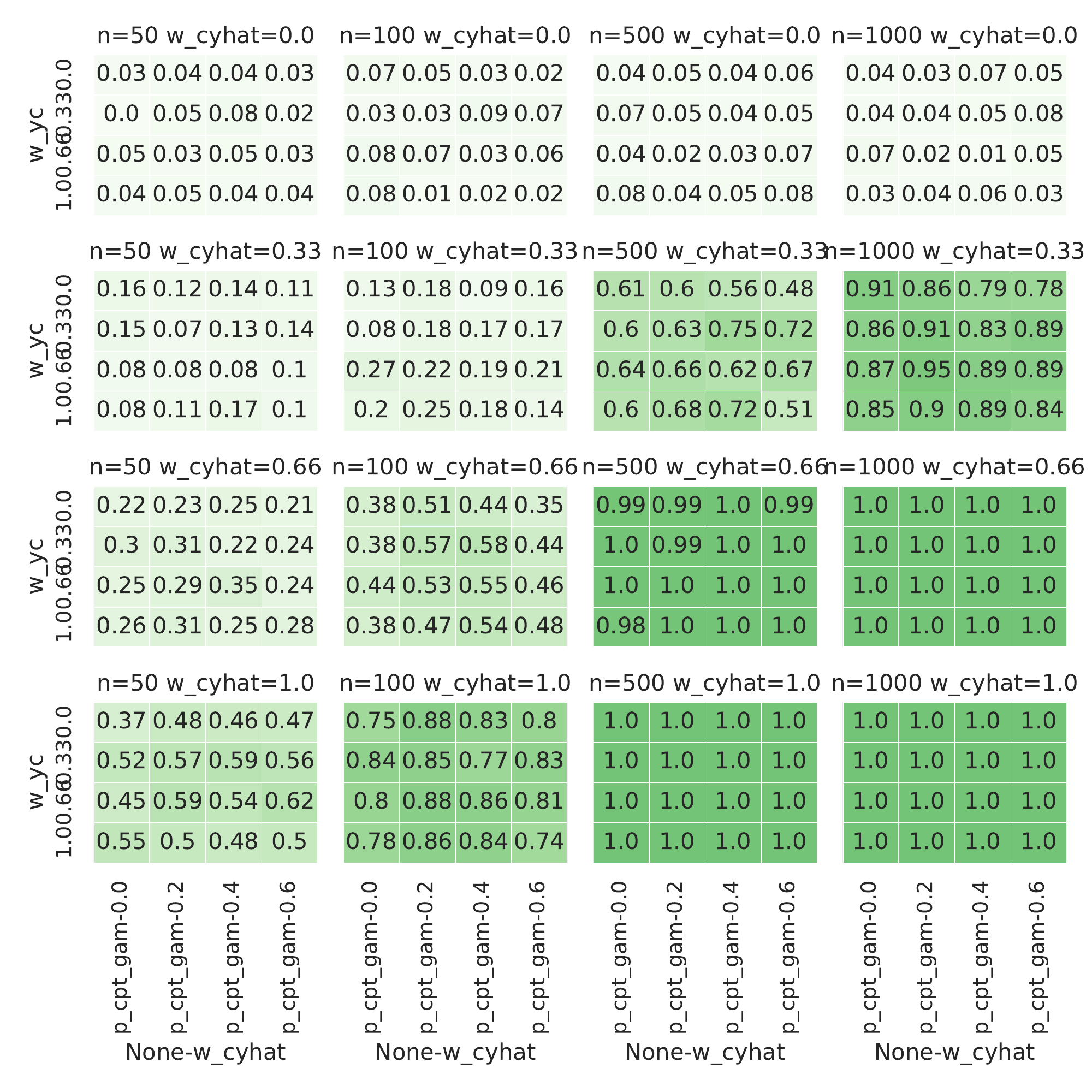}}
  \caption{Heatmaps showing the positive rates of the 'full' confounder test, with categorical variables, normal conditional distribution and sigmoid dependence.}
  \label{fig:sim-bbb-sig-full}
\end{figure}

Simulations show that non-normality has minimal effect on the power of the tests, except in case of extreme non-normality. (Fig. \ref{fig:sim-non-normal}). Simulations with sigmoid dependence resulted in an apparent loss of statistical power, however this is simply a consequence of the simulation methodology: with the same parameters, the sigmoid transformed confounder explains only approximately half the variance as compared to linear simulations.
Type I error control was valid in case of categorical variables and for the \emph{full} confounder test, as well. Power in these cases was also highly similar to the that of the partial confounder test with numerical variables. (Supplementary Figures \ref{fig:sim-bbb-lin-partial}, \ref{fig:sim-bbb-lin-full}, \ref{fig:sim-bbb-sig-partial} and \ref{fig:sim-bbb-sig-full}).

\subsection{Neuroimaging data}

To demonstrate the usefulness of the proposed tests in detecting various types of confounding bias, they have been deployed in two typical research scenarios - a regression and a classification problem - where confounder effects are known to hamper the development of biomedically valid predictive models. 

\subsubsection*{HCP dataset}

\begin{figure}[!b]
  \centering
  \includegraphics[width=0.75\paperwidth]{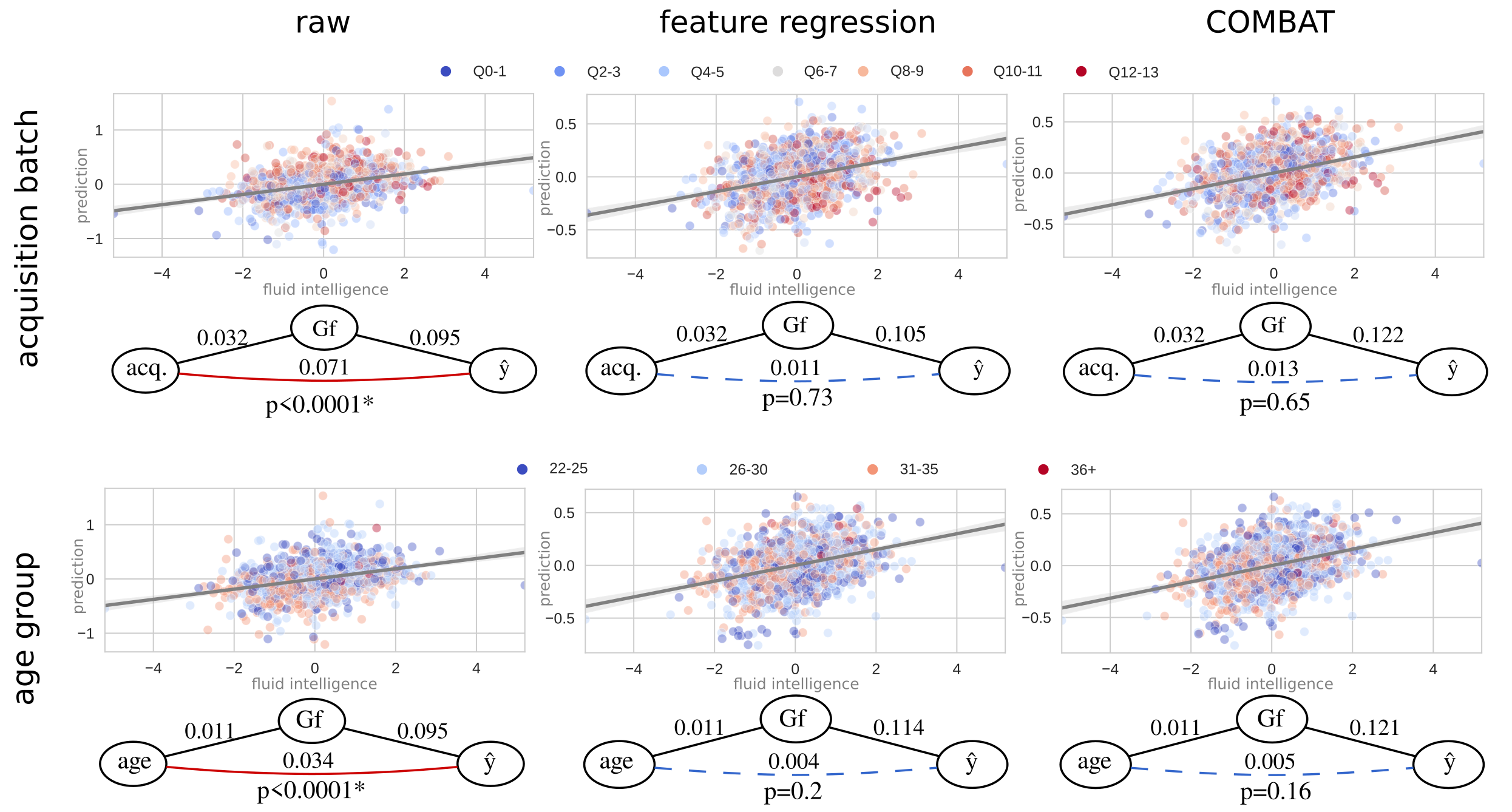}
  \caption{\textbf{Acquisition batch and age bias of fluid intelligence prediction in the HCP dataset.} \\
  Scatter plots and regression lines (with 95\% confidence intervals) show the association of the observed (horizontals axis) and predicted (vertical axis) fluid intelligence scores with various confound regression strategies. Color-coding of the confounder variables (top: acquisition batch, bottom: age group, as shown by the corresponding legends) reveals confounder bias both for acquisition and age in the models trained on the raw data. This bias is robustly detected by the partial confounder test ($p<0.0001$) and seems to be effectively mitigated by both feature regression and COMBAT.
  Relation between the observed ($Gf$) and predicted ($\hat{y}$) intelligence scores and the confounder variables is given on the graphs via $R^2 values$. Both confound mitigation techniques, but especially COMBAT, improve the predictive performance.
  Solid red line between the confounder and the prediction means significant confounding bias, whereas blue dashed line denotes that confounder testing provided no evidence for bias. P-values are determined with the partial confounder test. P-values of the 'full' confounder test (not shown) were all less then 0.0001, i.e. the confounders did not fully drive prediction for any of the models.
  }
  \label{fig:hcp}
\end{figure}

Functional connectivity data from the Human Connectome Project\citep{van2013wu} (HCP) was used to build predictive models of fluid intelligence ($G_f$) and to test for the previously discussed confounding effect of age\citep{lohmann2021predicting, dubois2018distributed} and,  additionally, the - somewhat underdiscussed - batch-like effect of acquisition date of the data within the course of the data acquisition process.

Both acquisition batch and age group were statistically significantly associated with $G_f$ ($R^2=0.032$ and $0.011$ and $p<0.001$ and $p=0.001$, respectively, see also Table \ref{tab:unconditional-pvals}). The model trained on the raw (unadjusted) connectivity features predicted fluid intelligence with a medium effect size ($R^2=0.095$, $p<0.001$).

The partial confounder test revealed that the 'raw' model (without confounder mitigation) was significantly biased both by age group and acquisition batch (both $p<0.0001$, first column of Fig. \ref{fig:hcp}) with later phases of the acquisition and lower age being associated to larger predicted values.

After applying confound mitigation approaches (feature regression or COMBAT) the partial confounder test did not provide evidence for confounding bias anymore ($p > 0.05$ for all; shown in the second and third columns of Fig. \ref{fig:hcp}), neither for acquisition batch nor for age. Both feature regression and COMBAT increased the predictive performance, with COMBAT providing the overall best performances ($R^2=0.122$ and $0.121$ when applied to remove the effect of acquisition and age, respectively).

The full confounder test was highly significant ($p<0.0001$) for all models, indicating that neither of the models were exclusively driven by the confounders.

\subsubsection*{ABIDE dataset}

\begin{figure}[!b]
  \centering
  \includegraphics[width=0.75\paperwidth]{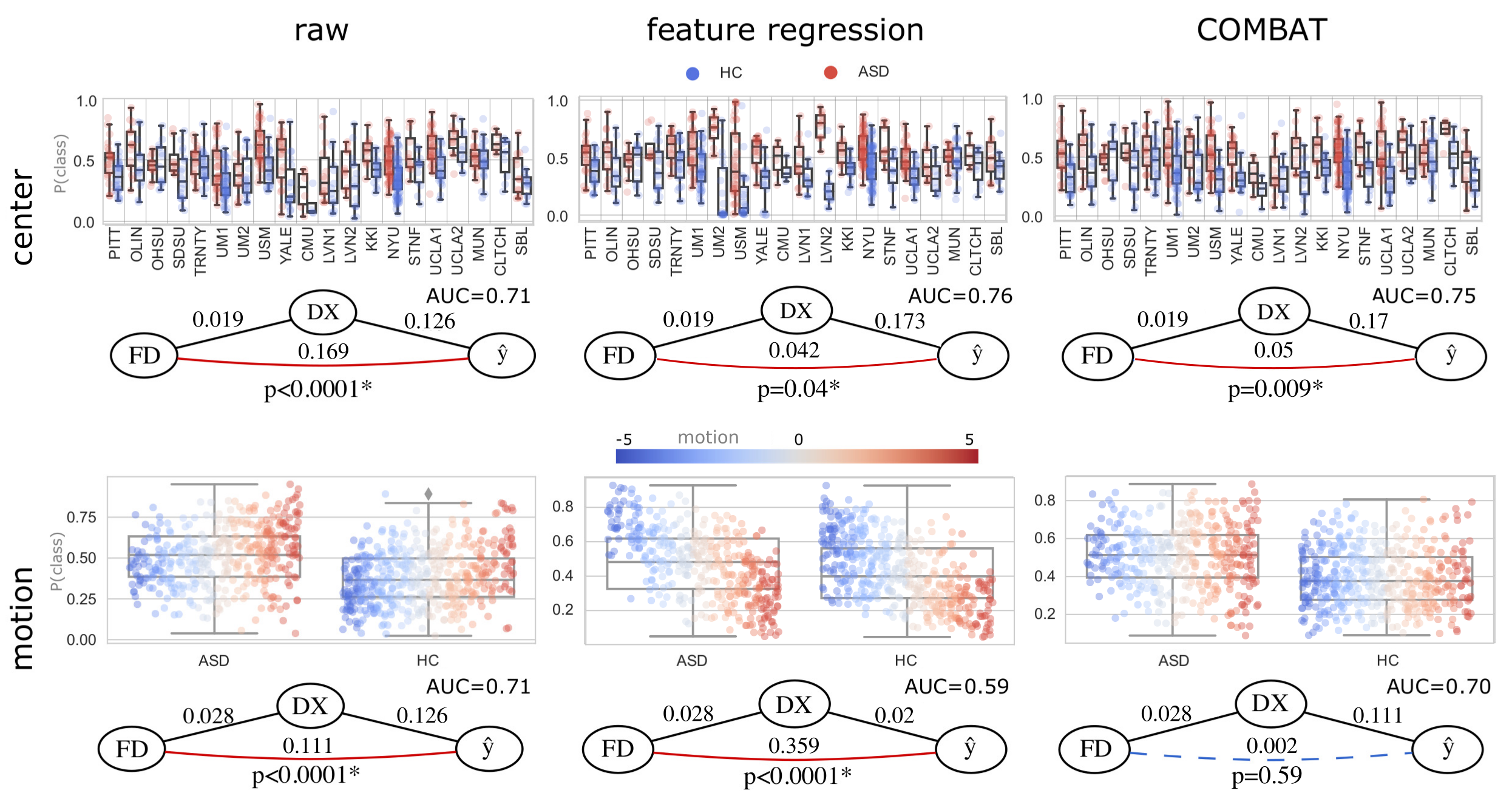}
  \caption{\textbf{Center- and motion age-bias of ASD diagnosis prediction in the ABIDE dataset.} \\
  Boxplots and points show the predicted class probabilities (0: HC, 1: ASD), separately for the HC and ASD groups. In the top panel, predictions are plotted for each center separately. Color indicates the true diagnosis (DX). AT the bottom plot, color indicates the normalized index of in-scanner motion (normalized FD). The proposed confounder test reveals significant center and motion bias in the model trained on the raw data (p<0.0001). While feature regression seems to effectively mitigate center bias, it fails to remove motion bias (actually introduces a paradoxical negative bias). COMBAT seems to effectively remove the effect of both confounders on the predictions and improves the predictive performance of the models in both cases.
  Relation between the true ($\hat{y}$) and predicted diagnosis scores and the confounder variables is shown by the graphs as $R^2$ values. Solid red line between the confounder and the prediction means significant confounding bias, whereas blue dashed line denotes that confounder testing provided no evidence for bias. P-values are determined with the partial confound test. P-values of the full confound test (not shown) were all less then 0.0001, i.e. the confounders did not fully drive the prediction for any of the models.
  }
  \label{fig:abide}
\end{figure}

Functional connectivity data from the ABIDE\citep{di2014autism} database was used to investigate the potential motion and center bias (as previously reported e.g. in refs\cite{spisak2014voxel, spisak2019optimal, gotts2013perils}) when training models that aim to predict ASD diagnosis.

Imaging center and in-scanner motion (normalized mean framewise displacement) were statistically significantly associated with ASD diagnosis ($R^2=0.019$ and $0.028$, respectively, $p<0.001$ for both, see also Table \ref{tab:unconditional-pvals}). The model trained on the raw (unadjusted) connectivity features predicted diagnosis with a medium effect size ($R^2=0.126$, $ROC AUC = 0.71$, $p<0.001$).

The partial confounder test revealed that the raw model was significantly biased both for age group and acquisition batch (both $p<0.0001$, see first column on Fig. \ref{fig:abide}). Predictions for several sites (e.g. Carnegine Mellon University, University of Leuven, Social Brain Lab UMC Groningen) were severely miscalibrated and higher motion was associated to a higher probability for ASD diagnosis.

Both feature regression and COMBAT seemed to significantly attenuate center bias, however the partial confound test still provided evidence for a significant bias ($p=0.04$ and $0.009$ for feature regression and COMBAT, respectively (second and third columns of the first row on Fig \ref{fig:abide}). 

When trying to mitigate the effect of in-scanner motion (bottom row on Fig \ref{fig:abide}), feature regression failed to remove the motion-bias from the model and, in fact, it introduced a paradoxical inverse dependence of the predictions on motion estimates. The partial confounder test successfully detected the resulting strong bias ($p<0.0001$). Applying COMBAT to remove the effect of motion (by using binned motion estimates), in turn, seemed to effectively mitigate motion-bias, as suggested by the partial confound test ($p=0.64$, bottom right panel of Fig \ref{fig:abide}).

Both feature regression and COMBAT considerably improved the predictive performance when mitigating center-effects ($R^2=0.111$ and $0.132$ and $AUC=0.76$ and $0.75$, respectively). With both feature regression and COMBAT, however, the effort of mitigating motion effects happened at the cost of a drop in predictive performance ($R^2=0.02$ and $0.111$ and $AUC=0.59$ and $0.70$, respectively)

The full confounder test was strongly significant ($p<0.0001$) for all models, except when regressing-out motion from the features ($p=0.09$), indicating that this model may be (almost) exclusively driven by motion artifacts.

\renewcommand{\arraystretch}{1.2}
\begin{table}
\centering
\begin{tabular}{lll|ll|ll|ll|ll} 
dataset & conf. & method & $R^2_{y, c}$ & $p_{y, c}$ & $R^2_{\hat{y}, c}$ & $p_{\hat{y}, c}$ & $R^2_{\hat{y}, y}$ & $p_{\hat{y}, y}$ & partial test & full test  \\
\hline
HCP & acq.  & raw      & 0.032 & <0.001 & 0.071 & <0.001  & 0.095 & <0.001 & \textbf{<0.0001} & <0.0001 \\
    &              & f.reg.    & & & 0.011 & 0.477 & 0.105  & <0.001  & 0.73 & <0.0001 \\
    &              & COMBAT    & & &0.013 & 0.4 & 0.122  & <0.001  & 0.65 & <0.0001\\
\hline
    & age   & raw       & 0.011 & 0.001  & 0.034 & <0.001  & 0.095  & <0.001 & \textbf{<0.0001} & <0.0001 \\
    &       & f.reg.    & & &0.004 & 0.052 & 0.114  & <0.001  & 0.2 & <0.0001 \\
    &       & COMBAT    & & &0.005 & 0.048 & 0.121 & <0.001 & 0.16 & <0.0001 \\
\hline
ABIDE   & center   & raw       & 0.019  & <0.001 &  0.169 & <0.001& 0.126     & <0.001 & \textbf{<0.0001} & <0.0001 \\
        &          & f.reg.    &  & &  0.042 & 0.007 & 0.173     & <0.001 & \textbf{0.04} & <0.0001 \\
        &          & COMBAT    &  & &  0.05 & 0.001 & 0.17     & <0.001 & \textbf{0.009} & <0.0001 \\
\hline
        & motion   & raw       & 0.028 & <0.001 & 0.111    &  <0.001 & 0.126    & <0.001 & \textbf{<0.0001} & <0.0001 \\
        &          & f.reg.    & & & 0.359    & <0.001  & 0.02   & <0.001 & \textbf{<0.0001} & \textbf{0.1} \\
        &          & COMBAT    & & &  0.002 & 0.19 & 0.111     & <0.001 & 0.59 & <0.0001 \\

\end{tabular}
\caption{\label{tab:unconditional-pvals} Coefficients-of-determination ($R^2$), the corresponding p-values and the p-values of the partial and full confounder tests, for all investigated datasets, confounders (conf.) and confounder-mitigation methods (method). Bold numbers denote significant confounding bias identified by the partial confounder test or the lack of significant contribution over that of the confounder, as identified by the full confounder test.}
\end{table}

\section{Discussion}

The assessment of confounding bias via conditional independence testing is a very appealing solution due to its simplicity and applicability in a wide variety of settings. However, statistical testing of conditional independence is challenging: 'conditional analogs' of bivariate (unconditioned) non-parametric tests can give invalid results even by slight violations of normality and linearity\citep{korn1984ranges} (see an example on Fig. \ref{fig:sim-h0-demo}). While the magnitude of this problem may not be fully appreciated in case of predictive model diagnostics, the 'no free lunch theorem' (i.e. the nonexistence of a uniformly valid conditional independence test)\cite{shah2020hardness} clearly implies that conditional independence-based confounding bias testing must be designed so that its suitability for the particular problem may be judged easily.

In this paper, I have proposed a solution for confounder testing, which was designed to be suitable for the specific characteristics of the 'confounder problem' in predictive modelling. Namely, the proposed approach places assumptions only about the conditional distribution of the confounder and the target variables, as required by the applied generalized additive\cite{hastie1987generalized} or multinomial logistic model\cite{bennett1966multiple, jones1975proability}. In biomedical research practice, these are variables that are often subject to descriptive (and often parametric) statistical analyses, as well. Accordingly, usually there is sufficient expert knowledge about the applicability of the required assumptions and the data acquisition is probably already designed so that the assumptions are likely fulfilled.
Of key importance, the proposed solution is \emph{absolutely assumption-free} regarding the distribution and dependency of the prediction output, which is, in most of the cases, unknown and can often be expected to be non-normal and/or non-linear (due to e.g model regularization)\citep{garcia2009study, kristensen2017whole}. 
This property distinguishes the approach from other alternatives as it guarantees a valid type-I error control even in cases of non-normally and non-linearly dependent predictions, i.e. in cases where Pearson and Spearman partial correlations, and many other methods fail.

The proposed approach gives rise to two different statistical tests for testing confounding bias: the \emph{full confounder test} probes whether the model's predictive performance can be attributed exclusively to the confounder and the \emph{partial confounder test} investigates whether the model utilizes any confounder-variance in the predictions, when controlled for the target variable. 
The tests can be applied for arbitrary classification or regression models, without having to re-fit the model, that is, with a negligible extra computational cost.

As expected from theory, both tests displayed a valid type I error control and a high, practically relevant statistical power in the simulations, even if both the predictions and the confounder are non-normally and/or non-linearly dependent on the target variable (except by extreme non-normality). This result confirms that the tests can be deployed in a wide variety of predictive modelling scenarios. While different biomedical applications may consider different amounts of bias to be relevant, the presented results can serve as a basis for power calculations, in order to identify the necessary sample size for proper model diagnostics.

A characteristic example for the potential areas of applications is the novel field of "predictive neuroscience", where applying predictive modelling and machine learning on functional neuroimaging data holds great potential for both revolutionizing our understanding of the physical basis of mind and delivering clinically useful tools for diagnostics or therapeutic decision making\citep{woo2017building, wager2013fmri, spisak2020pain}. However, the presence of confounders that are typical for biomedical research (e.g. sample demographics, center-effects) or specific to the data acquisition and processing approach (e.g. imaging artifacts) presents a great challenge to these efforts.
The usefulness of the proposed tests is demonstrated in two such examples, using the HCP\citep{van2013wu}  and the ABIDE\citep{di2014autism} datasets.

In case of the HCP dataset, the statistically significant age bias of the 'raw' model for predicting fluid intelligence is in line with previous findings\citep{dubois2018distributed, lohmann2021predicting} and could likely exaggerate to a serious bias when testing the model on data of participants outside of the - relatively narrow - age range of the HCP sample. In this case, the bias would likely significantly harm the out-of-sample generalizability of this model. The bias of the same model for acquisition batch can also be problematic, especially as it has not yet been thoroughly discussed in case of the HCP dataset. There can be manifold reasons for the observed acquisition bias. Fluid intelligence of the included participants might be, for instance, affected by a changing selection bias during participant recruitment (e.g. as a consequence of the human connectome project receiving an increasing degree of public interest during its course).

In the ABIDE dataset, neither the center bias nor the age bias is surprising in the case of the 'raw' model but both would be obviously severely problematic for a diagnostic biomarker candidate of ASD. For instance the model trained on the raw (unadjusted) features - depending on the calibration of the predicted class probabilities - might classify all participants form e.g. the CMU (Carnegie Mellon University) center as neurotypical control participants. Similarly, the models biased by motion - next to having questionable neuroscientific validity - might systematically fail in populations with a tendency for higher in-scanner motion (as known for many conditions, among others ADHD\citep{eloyan2012automated} or Alzheimer's disease \citep{rao2017predictive}).
 
 The partial confounder test provided quantitative, statistically rigorous metrics for assessing the effectiveness of the investigated confounder mitigation techniques. In the HCP data, it revealed that both the acquisition bias and the age bias was very effectively removed by both feature regression and COMBAT (p>0.05 for all). Given the high power of the test at the sample sizes of the HCP dataset ($N=999$), any remaining confounder-bias is most probably very safely negligible and well out of the range of practical relevance.
 
On the other hand, the partial confounder test also showed, that the performance of the investigated confound mitigation techniques is much less convincing in the ABIDE dataset. The center bias of the classification in this massively multi-center dataset was, although mitigated, but not successfully removed by feature regression and COMBAT. Determining the relevance of the remaining bias is out of the scope of this paper, however significant p-values of the partial confounder test generally suggest that these models require more effective data harmonization.

Motion bias in the ABIDE dataset was not detectable anymore after COMBAT, but remained highly significant after feature regression.
Interestingly, in this latter case, the association between the confounder and the predictions become even stronger than before but with an opposite sign. At the same time, the prediction performance substantially dropped (although still remained statistically significant). 
A possible explanation for this phenomenon is that in-scanner motion, as previously described\citep{fournier2010motor, anzulewicz2016toward}, has manifold links to ASD and, therefore, regressing out motion estimates from the connectivity features might eliminate a significant amount of signal-of-interest. This effect might be boosted by the previously reported 'residual motion artifacts' and their regional interactions\citep{spisak2014voxel} and raises caution for using feature regression to mitigate the effect of similar, complexly distributed confounders.
 
 The success of COMBAT in eliminating motion bias is also not to be taken without any objections. As COMBAT was originally developed for harmonizing effects of categorical variables (e.g. center or batch), its application for continuous confounder variables is not trivial. Inputting discretized versions of continuous variables into COMBAT might be sub-optimal and raises further questions e.g regarding the optimal number of bins used during the discretization.
 
 In sum, the application of the \emph{partial} confounder test on the real data examples suggests that confounding bias should be much more carefully investigated and reported in studies utilizing predictive modelling and machine learning as (i) variables as trivial as the date of the acquisition can cause significant confounding bias and (ii) in certain situations, a sufficient mitigation of confounder-bias requires more effective solutions than feature regression or COMBAT and (iii) in some cases confounder mitigation can - paradoxically - introduce more bias. The partial confounder test can be considered as a useful, objective benchmark to guide the search for a suitable confounder mitigation approach for every dataset.
 
 Regarding the \emph{full} confounder test; while the real data examples in this paper were not typical cases of its potential applications, it has still provided important insights into model biases. Namely, in the case of the aforementioned paradox motion-regression model in the ABIDE dataset, the full confounder test did not provide any evidence that the model captures any extra variance over that explained by the confounder ($p=0.1$). In other words, the test accepted the null hypothesis of full model bias, suggesting that this model may be, in fact, almost exclusively driven by motion artifacts.
This example highlights that the full confounder test may become useful in exploratory phases of model development, where models might be still severely biased by various confounders and the questions is whether there is any biomedically relevant signal captured by the model, at all.

\section{Conclusion}

The lack of rigorous statistical tests for confounder-bias significantly hampers the development of predictive models in many fields of research.

To fill this critical gap in predictive model development, here I proposed two novel tests, the \emph{partial} and the \emph{full confounder tests}, which probe the null hypotheses of 'no confounding bias' and 'full confounding bias', respectively. 
The tests are distinguished from alternative approaches by their robustness to non-normally and non-linearly dependent predictions, rendering them applicable with a wide variety of machine learning models.
The tests have, moreover, a minimal computational overhead, as re-fitting the model is not required.

As demonstrated on functional brain connectivity-based predictive models of fluid intelligence and autism spectrum disorder, the tests can guide the optimization of the confound mitigation strategy and allow quantitative statistical assessment of the robustness, generalizability and neurobiological validity of predictive models in biomedical research.
Given their simplicity, robustness, wide applicability, high statistical power and computationally effective implementation (available in the  python package \emph{mlconfound}\footnote{\href{https://mlconfound.readthedocs.io}{https://mlconfound.readthedocs.io}}), the partial and full confounder tests emerge as novel tools in the methodological arsenal of predictive modelling and may largely accelerate the development of clinically useful machine learning biomarkers.

\newpage
\section{Methods}
\label{sec:methods}

\subsection{Notation and Background}

In a predictive modelling setting, let $\y$ denote the target variable, $\yhat$ denote model output, i.e. the predictions for $\y$ and let $\c$ denote a variable which is considered as a confounder (schematic representation on Fig. \ref{fig:schematic}B). Let us assume, furthermore, that $\y$, $\yhat$ and $\c$ consist of independent and identically distributed (IID) data points $(y_i, \hat{y}_i, c_i) \in Y \times \hat{Y} \times C$ for $i=1, \dots , n$ so that $\y=(y_1, \dots ,y_n)$, $\yhat=(\hat{y}_1, \dots, \hat{y}_n)$ and $\c=(c_1, \dots, c_n)$. 

Depending on the research question, the direct influence of $\c$ on the model predictions $\yhat$ is subject to be kept at a negligible level or it is to be proven that, at least, the model is not completely driven by $\c$.

Obviously, a strong association between $\yhat$ and $\c$ may indicate that the model is biased; its predictions are driven by the confounder rather than information about the target variable.
Assessing the simple bivariate (unconditioned) dependence ($H0: \yhat \independent \c$) between $\yhat$ and $\c$ (or any of the $\y$, $\yhat$, $\c$ variables) is, however, insufficient for the proper characterization of confounder-bias in predictive modelling.
For instance, even if $\yhat \independent \c$ is false, $\yhat$ might be only marginally dependent on $\c$, due to the dependence of both on $\y$. In other words, if the target variable $\y$ displays a true association to the confounder variable $\c$, a model that is completely blind to $\c$ (i.e not confounded at all) might still provide outputs $\yhat$ that are significantly associated with $\c$.

\subsubsection*{Conditional independence for testing confounding bias}

Instead of focusing on the 'unconditioned' independence between the confounder and the predictions, we shall consider the \emph{conditional independence} between $\yhat$ and $\c$ given $\y$ (written as $\yhat \independent \c | \y$) which, by definition \citep{dawid1979conditional}, means that $\mathbb{P}(\yhat, \c|\y) = \mathbb{P}(\yhat|\y)\mathbb{P}(\c|\y)$. The statistical test with this null hypothesis will be referred to as the \emph{partial confounder test}.

Alternatively, one might also be interested in testing $\yhat \independent \y | \c$. We will refer to the corresponding test as the \emph{full confounder test}.

 Conditional independence - in its general form - lies at the heart of several fundamental concepts in statistics and  plays an increasingly important role in various fields of applied statistics, particularly in biomedical applications\citep{spirtes2000causation, fiedler2011mediation, peters2016causal, candes2016panning}. Recently, Shah and colleagues\cite{shah2020hardness} have raised important concerns regarding conditional independence testing.
 Their "no free lunch" theorem implies that, without placing some assumptions on the joint distribution of $(\y, \yhat, \c)$, conditional independence testing is effectively impossible. In other words, neither the full nor the partial confounder tests can be constructed so that - for all distributions - they provide a valid type I error control and, at the same time, a non-trivial statistical power.

This result stands in strong contrast to \emph{unconditional} independence testing - where permutation tests \citep{pitman1937significance, fisher1942189}, provide a general, distribution-free solution - and it has important implications for confounder testing in predictive modelling where the distribution of the model outputs (conditioned on the target variable) - depending on the applied machine learning model - is unknown and often non-normal and non-linear. One of the trivial candidates for the task, partial correlation, for instance assumes that all involved variables are multivariate Gaussian and - as to be shown below in a simulated example - even its Spearman-based variant is unable to tolerate relatively small deviations from normality and linearity.

Recently, Cand\`es et al.\cite{candes2016panning}, and, based on their work, Berret et al.\cite{berrett2020conditional}, have demonstrated that valid and powerful conditional independence tests can be constructed with inputting distributional information about only two (out of the three) variables.
Specifically, the conditional permutation test (CPT) of Berrett and colleagues samples from a non-uniform distribution over the set of possible permutations $\pi$ of one of the variables, based on its conditional distribution of the other variable. Thereby, it incorporates the information available about the conditional distribution of interest into the permutation-based inference in a statistically valid manner.

As many related papers, the work of Berrett et al. was formalized as a (semi-)supervised learning approach, where $X$ is a set of predictors (features), $Y$ is the target variable and $Z$ is a potential confounder. In this setting, testing the null hypothesis $X \independent Y | Z$ aims to determine, whether the features $X$ still affect $Y$, when controlling for $Z$.
For instance, in genome-wide association studies, CPT can be used to determine whether a particular genetic variant $X$ affects a response $Y$ such as disease status or some other phenotype, even after controlling for the rest of the genome, encoded in $Z$.

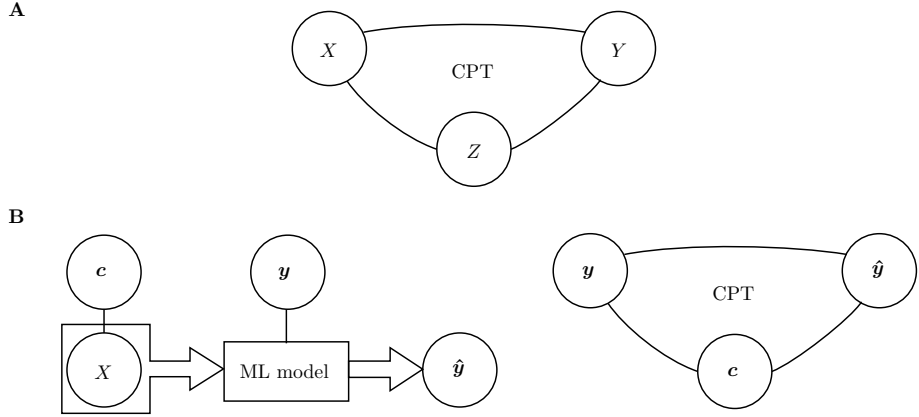
\begin{figure}
  \centering
  \resizebox{0.75\columnwidth}{!}{%
\tikzset{every picture/.style={line width=0.75pt}} 

\tikzset{every picture/.style={line width=0.75pt}} 

\begin{tikzpicture}[x=0.75pt,y=0.75pt,yscale=-1,xscale=1]

\draw   (299.5,107) .. controls (299.5,93.19) and (310.69,82) .. (324.5,82) .. controls (338.31,82) and (349.5,93.19) .. (349.5,107) .. controls (349.5,120.81) and (338.31,132) .. (324.5,132) .. controls (310.69,132) and (299.5,120.81) .. (299.5,107) -- cycle ;
\draw   (397,39) .. controls (397,25.19) and (408.19,14) .. (422,14) .. controls (435.81,14) and (447,25.19) .. (447,39) .. controls (447,52.81) and (435.81,64) .. (422,64) .. controls (408.19,64) and (397,52.81) .. (397,39) -- cycle ;
\draw   (202,39) .. controls (202,25.19) and (213.19,14) .. (227,14) .. controls (240.81,14) and (252,25.19) .. (252,39) .. controls (252,52.81) and (240.81,64) .. (227,64) .. controls (213.19,64) and (202,52.81) .. (202,39) -- cycle ;
\draw    (250,28) .. controls (292,20) and (364,22) .. (400,28) ;
\draw    (239,61) .. controls (249,76) and (276,99) .. (299.5,107) ;
\draw    (349.5,107) .. controls (365,101) and (403,71) .. (410,60) ;

\draw   (475.5,257) .. controls (475.5,243.19) and (486.69,232) .. (500.5,232) .. controls (514.31,232) and (525.5,243.19) .. (525.5,257) .. controls (525.5,270.81) and (514.31,282) .. (500.5,282) .. controls (486.69,282) and (475.5,270.81) .. (475.5,257) -- cycle ;
\draw   (573,189) .. controls (573,175.19) and (584.19,164) .. (598,164) .. controls (611.81,164) and (623,175.19) .. (623,189) .. controls (623,202.81) and (611.81,214) .. (598,214) .. controls (584.19,214) and (573,202.81) .. (573,189) -- cycle ;
\draw   (378,189) .. controls (378,175.19) and (389.19,164) .. (403,164) .. controls (416.81,164) and (428,175.19) .. (428,189) .. controls (428,202.81) and (416.81,214) .. (403,214) .. controls (389.19,214) and (378,202.81) .. (378,189) -- cycle ;
\draw    (426,178) .. controls (468,170) and (540,172) .. (576,178) ;
\draw    (415,211) .. controls (425,226) and (452,249) .. (475.5,257) ;
\draw    (525.5,257) .. controls (541,251) and (579,221) .. (586,210) ;
\draw   (50,256) .. controls (50,242.19) and (61.19,231) .. (75,231) .. controls (88.81,231) and (100,242.19) .. (100,256) .. controls (100,269.81) and (88.81,281) .. (75,281) .. controls (61.19,281) and (50,269.81) .. (50,256) -- cycle ;
\draw   (156,237) -- (240,237) -- (240,277) -- (156,277) -- cycle ;
\draw   (290,256) .. controls (290,242.19) and (301.19,231) .. (315,231) .. controls (328.81,231) and (340,242.19) .. (340,256) .. controls (340,269.81) and (328.81,281) .. (315,281) .. controls (301.19,281) and (290,269.81) .. (290,256) -- cycle ;
\draw   (174,190) .. controls (174,176.19) and (185.19,165) .. (199,165) .. controls (212.81,165) and (224,176.19) .. (224,190) .. controls (224,203.81) and (212.81,215) .. (199,215) .. controls (185.19,215) and (174,203.81) .. (174,190) -- cycle ;
\draw   (46.55,225.27) -- (106,225.27) -- (106,250) -- (133,250) -- (133,241) -- (155.55,255.27) -- (133,269.55) -- (133,260.55) -- (106,260.55) -- (106,285.27) -- (46.55,285.27) -- cycle ;
\draw   (240,250) -- (268,250) -- (268,241) -- (290,255.5) -- (268,270) -- (268,261) -- (240,261) -- cycle ;
\draw   (50,190) .. controls (50,176.19) and (61.19,165) .. (75,165) .. controls (88.81,165) and (100,176.19) .. (100,190) .. controls (100,203.81) and (88.81,215) .. (75,215) .. controls (61.19,215) and (50,203.81) .. (50,190) -- cycle ;
\draw    (75,215) -- (75,231) ;
\draw    (198,238) -- (198,215) ;

\draw (165,251) node [anchor=north west][inner sep=0.75pt]   [align=left] {ML model};
\draw (9,6) node [anchor=north west][inner sep=0.75pt]   [align=left] {\textbf{A}};
\draw (9,146) node [anchor=north west][inner sep=0.75pt]   [align=left] {\textbf{B}};
\draw (494,252.4) node [anchor=north west][inner sep=0.75pt]    {$\c$};
\draw (591.5,180.4) node [anchor=north west][inner sep=0.75pt]    {$\yhat$};
\draw (396,184.4) node [anchor=north west][inner sep=0.75pt]    {$\y$};
\draw (484,198) node [anchor=north west][inner sep=0.75pt]   [align=left] {CPT};
\draw (67,251.4) node [anchor=north west][inner sep=0.75pt]    {$X$};
\draw (308.5,247.4) node [anchor=north west][inner sep=0.75pt]    {$\yhat$};
\draw (191,185.4) node [anchor=north west][inner sep=0.75pt]    {$\y$};
\draw (318,102.4) node [anchor=north west][inner sep=0.75pt]    {$Z$};
\draw (415.5,34.4) node [anchor=north west][inner sep=0.75pt]    {$Y$};
\draw (219,34.4) node [anchor=north west][inner sep=0.75pt]    {$X$};
\draw (308,48) node [anchor=north west][inner sep=0.75pt]   [align=left] {CPT};
\draw (68,185.4) node [anchor=north west][inner sep=0.75pt]    {$\c$};

\end{tikzpicture}

    }
  \caption{Schematic diagram of using CPT in a predictive modelling context. \\ \textbf{(A)} CPT was originally proposed to be used on the feature variable $X$, target variable $Y$ and confounders $Z$. \textbf{(B)} The proposed use of CPT in predictive modelling requires the model to be fitted first, to obtain the model's prediction $\yhat$ on $\y$. CPT is then utilized on the triplet $(\y, \yhat, \c)$, to test hypotheses $\yhat \independent \c | \y$ or $\y \independent \yhat | \c$.}
  \label{fig:schematic}
\end{figure}

In this paper, a different setting is considered, where the supervised learning model is already fitted and we are focusing on model diagnostics, by testing the triplet $(\y,\yhat, \c)$. 
The conceptual difference between the original and the proposed application of CPT is depicted on Fig. \ref{fig:schematic}.

Within this setting, conditional independence testing and, specifically, the framework of conditional permutation testing allows investigating three different null hypotheses corresponding to the $(\y, \yhat, \c)$ triplet. As listed in Table \ref{tab:conditional-independence-cases}, testing the null hypothesis $\y \independent \yhat | \c$ (option 1, full confounder testing) investigates whether the model predictions are likely explainable solely with the confounder, i.e. whether the model is exclusively confounder-driven. Testing $\y \independent \c | \yhat$ (option 2) addresses the question whether the model captures all the variance in $c$ when predicting $y$. Testing the null hypothesis $\yhat \independent \c | \y$ (option 3, partial confounder testing) examines, whether the dependence of the model output on the confounder can likely be explained by the confounder's dependence on the target variable, i.e. whether there is any confounding bias in the model.

\renewcommand{\arraystretch}{1.2}
\begin{table}[]
\centering
\begin{tabular}{l|rp{60mm}|c|>{\centering\arraybackslash}m{30mm}}
 &  & H0  & assumption needed for: & no assumptions about the distribution of: \\
\hline
\textbf{1.} & $\yhat \independent \y | \c$ \quad  & full confounder test: model exclusively driven by the confounder & $Q(\y|\c)$ & $(\yhat,\y), (\yhat, \c)$ \\
\textbf{2.} & $\y \independent \c | \yhat$ \quad & model captures all variance in the confounder (not of interest) & $Q(\c|\yhat)$ & $(\y, \c), (\y, \yhat)$ \\
\textbf{3.} & $\yhat \independent \c | \y$  \quad &  partial confounder  test: model not directly driven by the confounder & $Q(\c|\y)$ & $(\yhat, \c), (\yhat, \y)$ \\
\end{tabular}
\caption{\label{tab:conditional-independence-cases} Possibilities when testing conditional independence in potentially biased predictive models. \\The table lists the three possible null hypotheses (H0), and the variables where assumption about the joint/conditional distributions is required/not required.   ($\y$: prediction target, $\yhat$: predictions, $\c$: confounder variable) }
\end{table}

Option 3, i.e. partial confounder testing is typically of interest when testing confounding bias of predictive models. Option 1, i.e. full confounder testing, may be also useful in diagnostics of predictive models, especially in the exploratory phase of model construction. Option 2 seems less appealing for model diagnostics and importantly, in this case the proposed variety of the CPT framework does not allow constructing a test which is non-parametric on $\yhat$.

In the following section, CPT is adapted for \emph{partial} confounder testing (option 3) and extended with general additive model\cite{hastie1987generalized} (GAM) and multinomial logistic regression\citep{bennett1966multiple, jones1975proability} based conditional distribution estimations, in order to make it handle categorical data and non-linear dependencies between the confounder and the target variable. (For an overview of the method, see Fig. \ref{fig:overview}). The formulation of the \emph{full} confounder test (option 1) is analogous and given in Supplementary Material \ref{sup:full-test}.

\subsubsection*{The partial confounder test}
\label{sec:methods-pct}

In accordance with the CPT approach, the partial confounder test generates a null-distribution for an arbitrary predefined test statistic $T(\y,\yhat,\c)$ by sampling permutation based 'copies' of $\c$,

\begin{equation}
    c_i^{(j)} \sim Q(\cdot|y_i)
     \label{eq:cond-copy}
\end{equation}

where, $Q(.|y)$ denotes the conditional distribution of $\c$ given $\y=y_i$ and $j=1,\dots, m$ indexes the 'copy' of $\c$ so that

$$ \c^{(j)} = (c_1^{(j)}, \dots, c_n^{(j)}) = (c_{\pi_1^{(j)}}, \dots, c_{\pi_n^{(j)}}) =  \c_{\boldsymbol{\pi}^{(j)}} $$

is a permutation of the original vector $\c = (c_1, \dots, c_n)$, with its elements reordered according to the permutation $ \boldsymbol{\pi} \in S_n$ where $S_n$ denote the set of all permutations on the indices $\{1,\dots,n\}$.

As shown by Berrett and colleagues\cite{berrett2020conditional}, to ensure that Eq. \ref{eq:cond-copy} holds, the $\c_{\boldsymbol{\pi}^{(j)}}$ copies must be drawn so that:

\begin{equation}
    \label{eq-pperm}
    \mathbb{P}(\boldsymbol{ \pi }^{(j)} = \boldsymbol{ \pi } | \y,\yhat,\c) = \frac{q^n(\c_{\boldsymbol{ \pi }} | \y)}{\sum_{\boldsymbol{ \pi } ' \in S_n} q^n(\c_{\boldsymbol{ \pi } ' } | \y)}
\end{equation}

that is, according to the $q^n(\cdot|\y) := q(\cdot | y_1) \dots q(\cdot|y_n)$ product density corresponding to the conditional distribution $Q(\cdot|\y)$. Note that Eq. \ref{eq-pperm} does not necessarily assume a continuous distribution.

This mechanism creates copies $\c^{(1)}, \dots ,\c^{(m)}$ so that under the null hypothesis ($\yhat \independent \c | \y$), the triples 

$$(\y,\yhat,\c), (\y, \yhat, \c^{(1)}),\dots, (\y, \yhat, \c^{(m)})$$

are all identically distributed and so are the 

$$T(\y,\yhat,\c), T(\y, \yhat, \c^{(1)}),\dots,T(\y, \yhat, \c^{(m)})$$

test statistics, as well.

As long as the numerator of Eq. \ref{eq-pperm} is non-zero for all $c_\pi \in C$ and $y \in Y$, the conditional permutations constitute an algebraic group, thus, as shown by Hemerik and Goeman\citep{hemerik2018exact}, an unbiased estimate of the p-value under the null can be obtained as:
$$ p= \frac{\sum_{j=1}^m \mathbb{1} \{T(\y, \yhat, \c^{(j)}) \geq T(\y, \yhat, \c) \}  }{m}$$

While the group property of the conditioned permutations provides a straightforward proof for the validity of the approach, for an alternative verification see the proof of Theorem 1 in Ref.\citep{berrett2020conditional}.

The required permutations could be theoretically sampled with a simple Metropolis-Hastings algorithm that draws uniformly from $S_n$ at random. However, this way the acceptance ratio would be extremely low, even for moderate $n$ (except there is very low dependence of $\c$ on $\y$), resulting in slow mixing times. The partial confounder test can be, however, efficiently implemented with the parallelized pairwise Markov-Chain Monte Carlo sampler of Berrett and colleagues\cite{berrett2020conditional} (Algorithm 1), that draws disjoint pairs in parallel and decides whether or not to swap them randomly, according to the odds ratio calculated from the conditional densities belonging to the original and swapped data. The acceptance odds ratio of swapping indices $i$ and $j$ is:

\begin{equation}
\frac{ q(c_j | y_i) q(c_i | y_j)}{q(c_i | y_i) q(c_j | y_j) }
=
\ell(c_j | y_i) + \ell(c_i | y_j) - \ell(c_i | y_i) - \ell(c_j | y_j) 
\label{eq:accept-odds}
\end{equation}

where $\ell$ denotes the log-likelihood.

In their Theorem 2, Berrett et al.\cite{berrett2020conditional} verify that the resulting Markov Chain yields the desired stationary distribution, even if the number of steps is small.

\subsubsection*{Conditional log-likelihood}

Obtaining a relatively accurate, independent estimate of $D(\cdot|\y)$ (of any shape) for CPT inference is important. Berrett and colleagues recommend to use a large independent sample to obtain the log-likelihood matrix that represents the conditional distribution $D(\cdot|Z)$ or, alternatively, to re-use the data by fitting a least squares linear regression:

\begin{equation}
    \label{eq:linreg}
    \c = \alpha + \beta \y + \boldsymbol{e}
\end{equation}

As the linear regression-based method, obviously, does not handle non-linear relationships I propose to fit a generalized additive model (GAM)\citep{hastie1987generalized}, instead:

\begin{equation}
    \label{eq:gam}
    \c = \alpha + \beta f(\y) + \boldsymbol{e}
\end{equation}

where the feature functions $f$ is built using penalized B-splines, which allow us to automatically model non-linear relationships without having to manually try out many different transformations on each variable.

Smoothness of the GAM model is optimized with a grid-search by picking the model with the lowest generalized cross-validation score form the models defined by the default parameters as implemented in PyGAM\citep{serven2018generalized} (v0.8.0).

If we write $\boldsymbol{\mu} = \alpha + \beta f(\y)$ and $\sigma$ denotes the standard deviation of the residual $\boldsymbol{e}$, then the conditional distribution of interest can be assumed to be normal with the parameters:

$$ (\c|\y=y_i) \sim \mathcal{N}\{\mu_i, \sigma^2\}$$

and the log-likelihood, that is to be used in Eq. \ref{eq:accept-odds}, can be computed simply as the log of the corresponding probability density function:

$$ \ell(c_i|y_j) = - \frac{1}{2} \Big(\frac{c_i-\mu_j}{\sigma}\Big)^2 - ln(2 \pi \sigma)   $$

In the case of categorical $\c$, a multinomial logistic regression (\emph{mnlogit}) model can be used to obtain $D(\cdot|\y)$, with the extra assumption of \emph{complete separation} if $\y$ is also categorical (in order to ensure an invertible Hessian, see e.g. Ref.\citep{bennett1966multiple, jones1975proability}).

Importantly, both the GAM- and the \emph{mnlogit}-based approaches guarantee that the numerator of Eq. \ref{eq-pperm} is always greater than zero and the group property for the permutations holds.

Note that from the three options for conditional independence-based null hypotheses enumerated in Table \ref{tab:conditional-independence-cases}, the proposed approach can not provide a test for option 2 that is assumption-free about $\yhat$, as the variable, on which the independence is conditional, must be always the predictor variable in Eq. \ref{eq:gam}. However, as discussed above, this option is of low practical relevance, anyway.
Pleasingly,  the proposed Gaussian regression-based conditional likelihood estimation ensures that no assumptions on $\yhat$ have to be made for the practically relevant options 1 and 3, i.e. for the full and partial confounder tests.

In theory, any predefined test statistic $T$ can be used with the proposed approach. The python package \emph{mlconfound}, implementing the proposed full and partial confounder tests, utilizes the coefficient of determination ($R^2$ or pseudo $R^2$ in case of categorical confounder or classification\citep{starkweather2011multinomial}) as a test statistic:

$$T(\y, \yhat, \c) = R^2(\yhat, \c)$$ 

and

$$T(\y, \yhat, \c^{(j)}) = R^2(\yhat, \c^{(j)})$$ 

which allows interpretable, two tailed inference about confounder-bias in predictive modelling.

\subsection{Validation on simulated data}

Using CPT to test confounding bias in predictive modelling allows relaxing assumptions on $\yhat$ (for both the partial and the full test) but - in line with the "no free lunch" theorem, requires knowing - or putting assumptions on - the joint distribution of the other two variables ($\y$ and $\c$). 
Berrett et al. \cite{berrett2020conditional} give a detailed analysis of the robustness of their CPT approach when estimating the conditional distribution with re-using the tested data via linear regression and, also, against misspecifying the conditional distribution of interest to introduce non-linearity.

Here I extend these results by performing simulations that evaluate the GAM- and \emph{mnlogit}-based approaches, in a form that is accessible for power calculations in predictive modelling (considering various weights of the target signal in $\c$ and the confounder and the target signals in $\yhat$).
Moreover, I investigate the robustness of the tests against the violation of normality and linearity of the conditional distributions $D(\c|\y)$ and $D(\yhat|\y)$.

Simulations are performed separately for the two proposed tests.

\subsubsection*{Simulation approach}
As  a first step, the target variable $\y$ is drawn randomly from a normal distribution:

$$ \y \sim \mathcal{N}(0, 1) $$

Next, the confounder signal is simulated as:

$$ \c | y_i \sim f_{\delta, \epsilon}(\mathcal{N}(0, 1)) + w_{yc} \, g(y_i) $$

where $f$ is a function to introduce non-normality, namely the \emph{sinh-arcsinh} transformation of Jones et al. \cite{jones2009sinh}, defined as:

$$f_{\delta, \epsilon}(\boldsymbol{x}) = sinh(\delta sinh^{-1}(\boldsymbol{x}) - \epsilon)$$

where the parameters $\delta$ and $\epsilon$ control the kurtosis and skewness of the resulting \emph{sinh-arcsinh} distribution, with $\delta=1$ and $\epsilon=0$ producing the identity function (i.e. no non-normality introduced).

Moreover, non-linearity can be introduced with the function $g$, which can be simply the identity function (no non-linearity is introduced in this case) or, for instance, a sigmoid-shaped function, in our case:

$$ g(\boldsymbol{x}) = tanh(\boldsymbol{x}) $$

The simulated predicted values are constructed in a similar fashion, but may depend on $\c$ as well:

$$ \yhat | y_i, c_i \sim f_{\delta, \epsilon}(\mathcal{N}(0, 1)) + w_{y\hat{y}} \: g(y_i) \, + \, w_{c\hat{y}} \: c_i$$

Note that simulations with $w_{c\hat{y}}=0$ produce data under the null hypothesis of no confounding bias.

To test the implementation for categorical variables, simulated $\y$, $\yhat$ and $\c$ variables are binarized by thresholding at 0.

\subsubsection*{Simulations for comparison with partial Spearman correlation and linear CPT}

To demonstrate the need for the proposed GAM-based CPT approach for partial confounder testing (Fig. \ref{fig:sim-h0-demo}), its validity was contrasted to partial Spearman correlation and the linear variety of CPT (based on eq. \ref{eq:linreg}, as described by\cite{berrett2020conditional}) with the following simulation parameters:
\begin{itemize}
    \item sample size: $n = 1000$
    \item $w_{c\hat{y}} = 0$ (i.e. H0 simulations only)
    \item all combinations of:
    \begin{itemize}
        \item $w_{yc} \in \{0.5, 1, 2, 3\}$ and
        \item $w_{y\hat{y}} \in \{0.5, 1, 2, 3\}$
    \end{itemize}
    \item non-normality ($f_{\delta = 0.1, \epsilon = 2}$) and non-linearity (sigmoid $g$)
\end{itemize}

For each parameter combination, 1000 repetitions were performed and false positive rates were calculated as the ratio of p-values smaller than $\alpha = 0.05$.

The simulation cases are exemplified (with $w_{yc} = w_{y\hat{y}} = 2$) on the left of Figure \ref{fig:sim-h0-demo}.

\subsubsection*{Simulations for evaluating power.}

Both for the partial and the full confounder test, 100 repetitions were performed of all combination of the following parameter values: 
\begin{itemize}
    \item $w_{yc} \in \{0.5, 1, 2, 3\}$
    \item $w_{y\hat{y}} \in \{0.5, 1, 2, 3\}$
    \item $w_{c\hat{y}} \in \{0, 0.2, 0.4, 0.6\}$
    \item $n \in \{50, 100, 500, 1000\}$
    \item linear and sigmoid dependence
    \item normal conditional distribution
\end{itemize}

The partial and full confound tests, as implemented in version 0.20.0 of the package \emph{'mlconfound'} were run with default parameters (1000 permutations and 50 Markov-chain Monte-Carlo steps to generate the conditioned permutations) and by implying categorical variables, where needed.

All code used for the simulations is available on github\footnote{\href{https://github.com/pni-lab/mlconfound-manuscript/tree/main/simulated}{https://github.com/pni-lab/mlconfound-manuscript/tree/main/simulated}}.

\subsection{Application on functional brain connectivity data}

The usefulness of the proposed confounder tests is demonstrated by applying them for predictive classification and regression models based on functional brain connectivity data, processed with different confound-mitigation approaches. 

Partial and full confounder testing was performed with 10000 permutations and 50 Markov-chain Monte Carlo steps, as implemented in version 0.20.0 of the package \emph{'mlconfound'}. Unconditional dependence across the involved variables was investigated with conventional permutation tests on the $R^2$ values, with 1000 permutations. 

All empirical analyses are available as jupyter notebooks on github\footnote{\href{https://github.com/pni-lab/mlconfound-manuscript/tree/main/empirical}{https://github.com/pni-lab/mlconfound-manuscript/tree/main/empirical}}.

\subsubsection*{HCP: testing age and acquisition batch bias in fluid intelligence prediction}

The Human Connectome Project dataset contains imaging and behavioral data of approximately 1200 healthy subjects\citep{van2013wu}. Preprocessed resting state fMRI connectivity data (partial correlation matrices)\citep{glasser2013minimal} as published with the HCP1200 release (N=999 participants with functional connectivity data) were used to build models that predict individual fluid intelligence scores ($G_f$), measured with Penn Progressive Matrices\citep{duncan2000neural}.

To ensure normality, $G_f$ was non-linearly transformed to normal distribution with the quantile transformation\citep{beasley2009rank} as implemented in \emph{scikit-learn}\citep{pedregosa2011scikit} (see Supplementary Figure \ref{fig:hcp-hist} for details).

Features (functional connectivities across 100 group-independent component analysis based regions) were either (i) considered in their raw form or were subject to confound mitigation approaches by (ii) feature regression\citep{rao2017predictive} or (iii) COMBAT\citep{johnson2007adjusting, fortin2018harmonization}.
The feature mitigation strategies were separately applied for acquisition batch and age group as confounder variable.

Each of the 5 types of features (raw, regressing out acquisition batch, regressing out age group, COMBAT with acquisition batch, COMBAT with age group) was independently incorporated into a scikit-learn-based\citep{pedregosa2011scikit} machine learning procedure aiming to predict the individual fluid intelligence scores with a ridge regression\citep{hoerl1970ridge}. The $\alpha$ parameter of the ridge model was considered as a hyperparameter ($\alpha \in \{0.00001, 0.0001, 0.001, 0.01, 0.1, 1, 10, 100, 1000, 10000, 100000\}$) and optimized in a nested cross-validation with 10 folds both in the inner and the outer loop and with mean squared error as optimization metric. Confound mitigation was performed inside of the outer cross-validation loop, to avoid leakage.

\subsubsection*{ABIDE: testing motion- and center-bias in predictive models of autism spectrum disorder diagnosis}

The proposed tests were applied to provide evidence of center- and  motion-bias in diagnostic predictive models of autism spectrum disorder (ASD), trained on the Autism Brain Imaging Data Exchange (ABIDE) dataset\citep{di2014autism} involving 866 participants (ASD: 402, neurotypical control: 464). Preprocessed regional timeseries data was obtained as shared with the by Dadi et al.\citep{dadi2019benchmarking} which was based on preprocessed image data provided by the Preprocessed Connectome Project\citep{craddock2013neuro}.

Tangent correlation across the timeseries of the n=122 regions of the BASC\citep{bellec2010multi} brain atlas was computed with nilearn\footnote{\href{http://nilearn.github.io/}{http://nilearn.github.io/}}\citep{huntenburg2017loading, esteve2015big}. 

The resulting functional connectivity estimates were considered as features either (i) in their raw form or after applying (ii) feature regression\citep{rao2017predictive} or (iii) COMBAT\citep{johnson2007adjusting, fortin2018harmonization}.
The investigated confounder variables were 'imaging center' and 'in-scanner motion', as measured by the mean framewise displacement (FD), as defined by Power and colleagues\cite{power2014methods}.
Mean FD was non-linearly transformed to normal distribution with the quantile transformation\citep{beasley2009rank} as implemented in \emph{scikit-learn}\citep{pedregosa2011scikit} (see Supplementary Figure \ref{fig:abide-hist} for details).

As COMBAT is not able to handle continuous variables (since it was primarily designed to remove categorical "batch-effects"), motion was binned into 10 groups, based on equidistant data quantiles ranging from 0 to 1.

The total of five types (raw, feature regression of site, feature regression of motion, COMBAT with site, COMBAT with motion) of features were independently incorporated into a scikit-learn-based\citep{pedregosa2011scikit} machine learning procedure aiming to predict the diagnosis (DX: ASD vs. neurotypical controls) with a L2-regularized logistic regression, as previously recommended\citep{dadi2019benchmarking}. The $C$ parameter of the model was considered as a hyperparameter ($C \in \{0.1, 1, 10\}$) and optimized in a nested cross-validation with 10 folds both in the inner and the outer cv-s and with area under the receiver operator curve (AUC under ROC) as optimization metric. Confound mitigation was performed inside of the outer cross-validation loop, to avoid leakage.
Confounder testing was performed on the predicted class probabilities.

\section{Acknowledgements}
I am thankful to Ulrike Bingel (University Hospital Essen, Germany) and Robert Englert (University Hospital Essen, Germany) for their valuable insights and comments on the manuscript. This research was supported by the Deutsche Forschungsgemeinschaft (DFG, German Research Foundation) – Projektnummer 316803389 – SFB 1280  and TRR 289 Treatment Expectation - Projektnummer 422744262.

\bibliographystyle{naturemag} 
\bibliography{references}

\begin{thebibliography}{10}
\expandafter\ifx\csname url\endcsname\relax
  \def\url#1{\texttt{#1}}\fi
\expandafter\ifx\csname urlprefix\endcsname\relax\def\urlprefix{URL }\fi
\providecommand{\bibinfo}[2]{#2}
\providecommand{\eprint}[2][]{\url{#2}}

\bibitem{vogt2018machine}
\bibinfo{author}{Vogt, N.}
\newblock \bibinfo{title}{Machine learning in neuroscience}.
\newblock \emph{\bibinfo{journal}{Nature Methods}}
  \textbf{\bibinfo{volume}{15}}, \bibinfo{pages}{33--33}
  (\bibinfo{year}{2018}).

\bibitem{kent2018personalized}
\bibinfo{author}{Kent, D.~M.}, \bibinfo{author}{Steyerberg, E.} \&
  \bibinfo{author}{van Klaveren, D.}
\newblock \bibinfo{title}{Personalized evidence based medicine: predictive
  approaches to heterogeneous treatment effects}.
\newblock \emph{\bibinfo{journal}{Bmj}} \textbf{\bibinfo{volume}{363}}
  (\bibinfo{year}{2018}).

\bibitem{spisak2020pain}
\bibinfo{author}{Spisak, T.} \emph{et~al.}
\newblock \bibinfo{title}{Pain-free resting-state functional brain connectivity
  predicts individual pain sensitivity}.
\newblock \emph{\bibinfo{journal}{Nature communications}}
  \textbf{\bibinfo{volume}{11}}, \bibinfo{pages}{1--12} (\bibinfo{year}{2020}).

\bibitem{walsh2021dome}
\bibinfo{author}{Walsh, I.} \emph{et~al.}
\newblock \bibinfo{title}{Dome: recommendations for supervised machine learning
  validation in biology}.
\newblock \emph{\bibinfo{journal}{Nature methods}} \bibinfo{pages}{1--6}
  (\bibinfo{year}{2021}).

\bibitem{woo2017building}
\bibinfo{author}{Woo, C.-W.}, \bibinfo{author}{Chang, L.~J.},
  \bibinfo{author}{Lindquist, M.~A.} \& \bibinfo{author}{Wager, T.~D.}
\newblock \bibinfo{title}{Building better biomarkers: brain models in
  translational neuroimaging}.
\newblock \emph{\bibinfo{journal}{Nature neuroscience}}
  \textbf{\bibinfo{volume}{20}}, \bibinfo{pages}{365--377}
  (\bibinfo{year}{2017}).

\bibitem{obermeyer2019dissecting}
\bibinfo{author}{Obermeyer, Z.}, \bibinfo{author}{Powers, B.},
  \bibinfo{author}{Vogeli, C.} \& \bibinfo{author}{Mullainathan, S.}
\newblock \bibinfo{title}{Dissecting racial bias in an algorithm used to manage
  the health of populations}.
\newblock \emph{\bibinfo{journal}{Science}} \textbf{\bibinfo{volume}{366}},
  \bibinfo{pages}{447--453} (\bibinfo{year}{2019}).

\bibitem{mehrabi2021survey}
\bibinfo{author}{Mehrabi, N.}, \bibinfo{author}{Morstatter, F.},
  \bibinfo{author}{Saxena, N.}, \bibinfo{author}{Lerman, K.} \&
  \bibinfo{author}{Galstyan, A.}
\newblock \bibinfo{title}{A survey on bias and fairness in machine learning}.
\newblock \emph{\bibinfo{journal}{ACM Computing Surveys (CSUR)}}
  \textbf{\bibinfo{volume}{54}}, \bibinfo{pages}{1--35} (\bibinfo{year}{2021}).

\bibitem{prosperi2020causal}
\bibinfo{author}{Prosperi, M.} \emph{et~al.}
\newblock \bibinfo{title}{Causal inference and counterfactual prediction in
  machine learning for actionable healthcare}.
\newblock \emph{\bibinfo{journal}{Nature Machine Intelligence}}
  \textbf{\bibinfo{volume}{2}}, \bibinfo{pages}{369--375}
  (\bibinfo{year}{2020}).

\bibitem{rao2017predictive}
\bibinfo{author}{Rao, A.}, \bibinfo{author}{Monteiro, J.~M.},
  \bibinfo{author}{Mourao-Miranda, J.}, \bibinfo{author}{Initiative, A.~D.}
  \emph{et~al.}
\newblock \bibinfo{title}{Predictive modelling using neuroimaging data in the
  presence of confounds}.
\newblock \emph{\bibinfo{journal}{NeuroImage}} \textbf{\bibinfo{volume}{150}},
  \bibinfo{pages}{23--49} (\bibinfo{year}{2017}).

\bibitem{eloyan2012automated}
\bibinfo{author}{Eloyan, A.} \emph{et~al.}
\newblock \bibinfo{title}{Automated diagnoses of attention deficit hyperactive
  disorder using magnetic resonance imaging}.
\newblock \emph{\bibinfo{journal}{Frontiers in systems neuroscience}}
  \textbf{\bibinfo{volume}{6}}, \bibinfo{pages}{61} (\bibinfo{year}{2012}).

\bibitem{couvy2016head}
\bibinfo{author}{Couvy-Duchesne, B.} \emph{et~al.}
\newblock \bibinfo{title}{Head motion and inattention/hyperactivity share
  common genetic influences: implications for fmri studies of adhd}.
\newblock \emph{\bibinfo{journal}{PloS one}} \textbf{\bibinfo{volume}{11}},
  \bibinfo{pages}{e0146271} (\bibinfo{year}{2016}).

\bibitem{gotts2013perils}
\bibinfo{author}{Gotts, S.~J.} \emph{et~al.}
\newblock \bibinfo{title}{The perils of global signal regression for group
  comparisons: a case study of autism spectrum disorders}.
\newblock \emph{\bibinfo{journal}{Frontiers in human neuroscience}}
  \textbf{\bibinfo{volume}{7}}, \bibinfo{pages}{356} (\bibinfo{year}{2013}).

\bibitem{spisak2014voxel}
\bibinfo{author}{Spisak, T.} \emph{et~al.}
\newblock \bibinfo{title}{Voxel-wise motion artifacts in population-level
  whole-brain connectivity analysis of resting-state fmri}.
\newblock \emph{\bibinfo{journal}{PloS one}} \textbf{\bibinfo{volume}{9}},
  \bibinfo{pages}{e104947} (\bibinfo{year}{2014}).

\bibitem{spisak2019optimal}
\bibinfo{author}{Spisak, T.}, \bibinfo{author}{Kincses, B.} \&
  \bibinfo{author}{Bingel, U.}
\newblock \bibinfo{title}{Optimal choice of parameters in functional
  connectome-based predictive modelling might be biased by motion: comment on
  dadi et al}.
\newblock \emph{\bibinfo{journal}{bioRxiv}} \bibinfo{pages}{710731}
  (\bibinfo{year}{2019}).

\bibitem{cole2012global}
\bibinfo{author}{Cole, M.~W.}, \bibinfo{author}{Yarkoni, T.},
  \bibinfo{author}{Repov{\v{s}}, G.}, \bibinfo{author}{Anticevic, A.} \&
  \bibinfo{author}{Braver, T.~S.}
\newblock \bibinfo{title}{Global connectivity of prefrontal cortex predicts
  cognitive control and intelligence}.
\newblock \emph{\bibinfo{journal}{Journal of Neuroscience}}
  \textbf{\bibinfo{volume}{32}}, \bibinfo{pages}{8988--8999}
  (\bibinfo{year}{2012}).

\bibitem{he2020deep}
\bibinfo{author}{He, T.} \emph{et~al.}
\newblock \bibinfo{title}{Deep neural networks and kernel regression achieve
  comparable accuracies for functional connectivity prediction of behavior and
  demographics}.
\newblock \emph{\bibinfo{journal}{NeuroImage}} \textbf{\bibinfo{volume}{206}},
  \bibinfo{pages}{116276} (\bibinfo{year}{2020}).

\bibitem{dubois2018distributed}
\bibinfo{author}{Dubois, J.}, \bibinfo{author}{Galdi, P.},
  \bibinfo{author}{Paul, L.~K.} \& \bibinfo{author}{Adolphs, R.}
\newblock \bibinfo{title}{A distributed brain network predicts general
  intelligence from resting-state human neuroimaging data}.
\newblock \emph{\bibinfo{journal}{Philosophical Transactions of the Royal
  Society B: Biological Sciences}} \textbf{\bibinfo{volume}{373}},
  \bibinfo{pages}{20170284} (\bibinfo{year}{2018}).

\bibitem{lohmann2021predicting}
\bibinfo{author}{Lohmann, G.} \emph{et~al.}
\newblock \bibinfo{title}{Predicting intelligence from fmri data of the human
  brain in a few minutes of scan time}.
\newblock \emph{\bibinfo{journal}{bioRxiv}}  (\bibinfo{year}{2021}).

\bibitem{lwowski2021risk}
\bibinfo{author}{Lwowski, B.} \& \bibinfo{author}{Rios, A.}
\newblock \bibinfo{title}{The risk of racial bias while tracking
  influenza-related content on social media using machine learning}.
\newblock \emph{\bibinfo{journal}{Journal of the American Medical Informatics
  Association}} \textbf{\bibinfo{volume}{28}}, \bibinfo{pages}{839--849}
  (\bibinfo{year}{2021}).

\bibitem{dukart2011age}
\bibinfo{author}{Dukart, J.}, \bibinfo{author}{Schroeter, M.~L.},
  \bibinfo{author}{Mueller, K.} \& \bibinfo{author}{Initiative, A. D.~N.}
\newblock \bibinfo{title}{Age correction in dementia--matching to a healthy
  brain}.
\newblock \emph{\bibinfo{journal}{PloS one}} \textbf{\bibinfo{volume}{6}},
  \bibinfo{pages}{e22193} (\bibinfo{year}{2011}).

\bibitem{abdulkadir2014reduction}
\bibinfo{author}{Abdulkadir, A.}, \bibinfo{author}{Ronneberger, O.},
  \bibinfo{author}{Tabrizi, S.~J.} \& \bibinfo{author}{Kl{\"o}ppel, S.}
\newblock \bibinfo{title}{Reduction of confounding effects with voxel-wise
  gaussian process regression in structural mri}.
\newblock In \emph{\bibinfo{booktitle}{2014 International Workshop on Pattern
  Recognition in Neuroimaging}}, \bibinfo{pages}{1--4}
  (\bibinfo{organization}{IEEE}, \bibinfo{year}{2014}).

\bibitem{johnson2007adjusting}
\bibinfo{author}{Johnson, W.~E.}, \bibinfo{author}{Li, C.} \&
  \bibinfo{author}{Rabinovic, A.}
\newblock \bibinfo{title}{Adjusting batch effects in microarray expression data
  using empirical bayes methods}.
\newblock \emph{\bibinfo{journal}{Biostatistics}} \textbf{\bibinfo{volume}{8}},
  \bibinfo{pages}{118--127} (\bibinfo{year}{2007}).

\bibitem{wachinger2021detect}
\bibinfo{author}{Wachinger, C.}, \bibinfo{author}{Rieckmann, A.},
  \bibinfo{author}{P{\"o}lsterl, S.}, \bibinfo{author}{Initiative, A. D.~N.}
  \emph{et~al.}
\newblock \bibinfo{title}{Detect and correct bias in multi-site neuroimaging
  datasets}.
\newblock \emph{\bibinfo{journal}{Medical Image Analysis}}
  \textbf{\bibinfo{volume}{67}}, \bibinfo{pages}{101879}
  (\bibinfo{year}{2021}).

\bibitem{korn1984ranges}
\bibinfo{author}{Korn, E.~L.}
\newblock \bibinfo{title}{The ranges of limiting values of some partial
  correlations under conditional independence}.
\newblock \emph{\bibinfo{journal}{The American Statistician}}
  \textbf{\bibinfo{volume}{38}}, \bibinfo{pages}{61--62}
  (\bibinfo{year}{1984}).

\bibitem{bergsma2010nonparametric}
\bibinfo{author}{Bergsma, W.}
\newblock \bibinfo{title}{Nonparametric testing of conditional independence by
  means of the partial copula}.
\newblock \emph{\bibinfo{journal}{Available at SSRN 1702981}}
  (\bibinfo{year}{2010}).

\bibitem{candes2016panning}
\bibinfo{author}{Cand\`es, E.}, \bibinfo{author}{Fan, Y.},
  \bibinfo{author}{Janson, L.} \& \bibinfo{author}{Lv, J.}
\newblock \bibinfo{title}{Panning for gold: Model-x knockoffs for
  high-dimensional controlled variable selection}.
\newblock \emph{\bibinfo{journal}{arXiv preprint arXiv:1610.02351}}
  (\bibinfo{year}{2016}).

\bibitem{peters2016causal}
\bibinfo{author}{Peters, J.}, \bibinfo{author}{B{\"u}hlmann, P.} \&
  \bibinfo{author}{Meinshausen, N.}
\newblock \bibinfo{title}{Causal inference by using invariant prediction:
  identification and confidence intervals}.
\newblock \emph{\bibinfo{journal}{Journal of the Royal Statistical Society.
  Series B (Statistical Methodology)}} \bibinfo{pages}{947--1012}
  (\bibinfo{year}{2016}).

\bibitem{shah2020hardness}
\bibinfo{author}{Shah, R.~D.} \& \bibinfo{author}{Peters, J.}
\newblock \bibinfo{title}{The hardness of conditional independence testing and
  the generalised covariance measure}.
\newblock \emph{\bibinfo{journal}{The Annals of Statistics}}
  \textbf{\bibinfo{volume}{48}}, \bibinfo{pages}{1514--1538}
  (\bibinfo{year}{2020}).

\bibitem{berrett2020conditional}
\bibinfo{author}{Berrett, T.~B.}, \bibinfo{author}{Wang, Y.},
  \bibinfo{author}{Barber, R.~F.} \& \bibinfo{author}{Samworth, R.~J.}
\newblock \bibinfo{title}{The conditional permutation test for independence
  while controlling for confounders}.
\newblock \emph{\bibinfo{journal}{Journal of the Royal Statistical Society:
  Series B (Statistical Methodology)}} \textbf{\bibinfo{volume}{82}},
  \bibinfo{pages}{175--197} (\bibinfo{year}{2020}).

\bibitem{garcia2009study}
\bibinfo{author}{Garc{\'\i}a, S.}, \bibinfo{author}{Fern{\'a}ndez, A.},
  \bibinfo{author}{Luengo, J.} \& \bibinfo{author}{Herrera, F.}
\newblock \bibinfo{title}{A study of statistical techniques and performance
  measures for genetics-based machine learning: accuracy and interpretability}.
\newblock \emph{\bibinfo{journal}{Soft Computing}}
  \textbf{\bibinfo{volume}{13}}, \bibinfo{pages}{959} (\bibinfo{year}{2009}).

\bibitem{kristensen2017whole}
\bibinfo{author}{Kristensen, S.~B.} \& \bibinfo{author}{Sandberg, K.}
\newblock \bibinfo{title}{Is whole-brain functional connectivity a neuromarker
  of sustained attention? comment on rosenberg \& al.(2016)}.
\newblock \emph{\bibinfo{journal}{bioRxiv}} \bibinfo{pages}{216697}
  (\bibinfo{year}{2017}).

\bibitem{chaibub2019permutation}
\bibinfo{author}{Neto, C.~E.} \emph{et~al.}
\newblock \bibinfo{title}{A permutation approach to assess confounding in
  machine learning applications for digital health}.
\newblock In \emph{\bibinfo{booktitle}{Proceedings of the 25th ACM SIGKDD
  International Conference on Knowledge Discovery \& Data Mining}},
  \bibinfo{pages}{54--64} (\bibinfo{year}{2019}).

\bibitem{ferrari2020measuring}
\bibinfo{author}{Ferrari, E.}, \bibinfo{author}{Retico, A.} \&
  \bibinfo{author}{Bacciu, D.}
\newblock \bibinfo{title}{Measuring the effects of confounders in medical
  supervised classification problems: the confounding index (ci)}.
\newblock \emph{\bibinfo{journal}{Artificial intelligence in medicine}}
  \textbf{\bibinfo{volume}{103}}, \bibinfo{pages}{101804}
  (\bibinfo{year}{2020}).

\bibitem{southworth2009properties}
\bibinfo{author}{Southworth, L.~K.}, \bibinfo{author}{Kim, S.~K.} \&
  \bibinfo{author}{Owen, A.~B.}
\newblock \bibinfo{title}{Properties of balanced permutations}.
\newblock \emph{\bibinfo{journal}{Journal of Computational Biology}}
  \textbf{\bibinfo{volume}{16}}, \bibinfo{pages}{625--638}
  (\bibinfo{year}{2009}).

\bibitem{hemerik2018exact}
\bibinfo{author}{Hemerik, J.} \& \bibinfo{author}{Goeman, J.}
\newblock \bibinfo{title}{Exact testing with random permutations}.
\newblock \emph{\bibinfo{journal}{Test}} \textbf{\bibinfo{volume}{27}},
  \bibinfo{pages}{811--825} (\bibinfo{year}{2018}).

\bibitem{hastie1987generalized}
\bibinfo{author}{Hastie, T.} \& \bibinfo{author}{Tibshirani, R.}
\newblock \bibinfo{title}{Generalized additive models: some applications}.
\newblock \emph{\bibinfo{journal}{Journal of the American Statistical
  Association}} \textbf{\bibinfo{volume}{82}}, \bibinfo{pages}{371--386}
  (\bibinfo{year}{1987}).

\bibitem{bennett1966multiple}
\bibinfo{author}{Bennett, B.}
\newblock \bibinfo{title}{Multiple regression analysis of binary and
  multinomial variates}.
\newblock \emph{\bibinfo{journal}{Sankhy{\=a}: The Indian Journal of
  Statistics, Series A}} \bibinfo{pages}{301--304} (\bibinfo{year}{1966}).

\bibitem{jones1975proability}
\bibinfo{author}{Jones, R.~H.}
\newblock \bibinfo{title}{Proability estimation usind a multinominal logistic
  function}.
\newblock \emph{\bibinfo{journal}{Journal of Statistical Computation and
  Simulation}} \textbf{\bibinfo{volume}{3}}, \bibinfo{pages}{315--329}
  (\bibinfo{year}{1975}).

\bibitem{van2013wu}
\bibinfo{author}{Van~Essen, D.~C.} \emph{et~al.}
\newblock \bibinfo{title}{The wu-minn human connectome project: an overview}.
\newblock \emph{\bibinfo{journal}{Neuroimage}} \textbf{\bibinfo{volume}{80}},
  \bibinfo{pages}{62--79} (\bibinfo{year}{2013}).

\bibitem{di2014autism}
\bibinfo{author}{Di~Martino, A.} \emph{et~al.}
\newblock \bibinfo{title}{The autism brain imaging data exchange: towards a
  large-scale evaluation of the intrinsic brain architecture in autism}.
\newblock \emph{\bibinfo{journal}{Molecular psychiatry}}
  \textbf{\bibinfo{volume}{19}}, \bibinfo{pages}{659--667}
  (\bibinfo{year}{2014}).

\bibitem{wager2013fmri}
\bibinfo{author}{Wager, T.~D.} \emph{et~al.}
\newblock \bibinfo{title}{An fmri-based neurologic signature of physical pain}.
\newblock \emph{\bibinfo{journal}{New England Journal of Medicine}}
  \textbf{\bibinfo{volume}{368}}, \bibinfo{pages}{1388--1397}
  (\bibinfo{year}{2013}).

\bibitem{fournier2010motor}
\bibinfo{author}{Fournier, K.~A.}, \bibinfo{author}{Hass, C.~J.},
  \bibinfo{author}{Naik, S.~K.}, \bibinfo{author}{Lodha, N.} \&
  \bibinfo{author}{Cauraugh, J.~H.}
\newblock \bibinfo{title}{Motor coordination in autism spectrum disorders: a
  synthesis and meta-analysis}.
\newblock \emph{\bibinfo{journal}{Journal of autism and developmental
  disorders}} \textbf{\bibinfo{volume}{40}}, \bibinfo{pages}{1227--1240}
  (\bibinfo{year}{2010}).

\bibitem{anzulewicz2016toward}
\bibinfo{author}{Anzulewicz, A.}, \bibinfo{author}{Sobota, K.} \&
  \bibinfo{author}{Delafield-Butt, J.~T.}
\newblock \bibinfo{title}{Toward the autism motor signature: Gesture patterns
  during smart tablet gameplay identify children with autism}.
\newblock \emph{\bibinfo{journal}{Scientific reports}}
  \textbf{\bibinfo{volume}{6}}, \bibinfo{pages}{1--13} (\bibinfo{year}{2016}).

\bibitem{dawid1979conditional}
\bibinfo{author}{Dawid, A.~P.}
\newblock \bibinfo{title}{Conditional independence in statistical theory}.
\newblock \emph{\bibinfo{journal}{Journal of the Royal Statistical Society:
  Series B (Methodological)}} \textbf{\bibinfo{volume}{41}},
  \bibinfo{pages}{1--15} (\bibinfo{year}{1979}).

\bibitem{spirtes2000causation}
\bibinfo{author}{Spirtes, P.}, \bibinfo{author}{Glymour, C.~N.},
  \bibinfo{author}{Scheines, R.} \& \bibinfo{author}{Heckerman, D.}
\newblock \emph{\bibinfo{title}{Causation, prediction, and search}}
  (\bibinfo{publisher}{MIT press}, \bibinfo{year}{2000}).

\bibitem{fiedler2011mediation}
\bibinfo{author}{Fiedler, K.}, \bibinfo{author}{Schott, M.} \&
  \bibinfo{author}{Meiser, T.}
\newblock \bibinfo{title}{What mediation analysis can (not) do}.
\newblock \emph{\bibinfo{journal}{Journal of Experimental Social Psychology}}
  \textbf{\bibinfo{volume}{47}}, \bibinfo{pages}{1231--1236}
  (\bibinfo{year}{2011}).

\bibitem{pitman1937significance}
\bibinfo{author}{Pitman, E.~J.}
\newblock \bibinfo{title}{Significance tests which may be applied to samples
  from any populations}.
\newblock \emph{\bibinfo{journal}{Supplement to the Journal of the Royal
  Statistical Society}} \textbf{\bibinfo{volume}{4}}, \bibinfo{pages}{119--130}
  (\bibinfo{year}{1937}).

\bibitem{fisher1942189}
\bibinfo{author}{Fisher, R.~A.} \emph{et~al.}
\newblock \bibinfo{title}{189: The theory of confounding in factorial
  experiments in relation to the theory of groups.}  (\bibinfo{year}{1942}).

\bibitem{serven2018generalized}
\bibinfo{author}{Servén, D.}, \bibinfo{author}{Brummitt, C.} \&
  \bibinfo{author}{Abedi, H.}
\newblock \bibinfo{title}{pygam: Generalized additive models in python}.
\newblock \emph{\bibinfo{journal}{Zenodo}}  (\bibinfo{year}{2018}).

\bibitem{starkweather2011multinomial}
\bibinfo{author}{Starkweather, J.} \& \bibinfo{author}{Moske, A.~K.}
\newblock \bibinfo{title}{Multinomial logistic regression}
  (\bibinfo{year}{2011}).

\bibitem{jones2009sinh}
\bibinfo{author}{Jones, M.~C.} \& \bibinfo{author}{Pewsey, A.}
\newblock \bibinfo{title}{Sinh-arcsinh distributions}.
\newblock \emph{\bibinfo{journal}{Biometrika}} \textbf{\bibinfo{volume}{96}},
  \bibinfo{pages}{761--780} (\bibinfo{year}{2009}).

\bibitem{glasser2013minimal}
\bibinfo{author}{Glasser, M.~F.} \emph{et~al.}
\newblock \bibinfo{title}{The minimal preprocessing pipelines for the human
  connectome project}.
\newblock \emph{\bibinfo{journal}{Neuroimage}} \textbf{\bibinfo{volume}{80}},
  \bibinfo{pages}{105--124} (\bibinfo{year}{2013}).

\bibitem{duncan2000neural}
\bibinfo{author}{Duncan, J.} \emph{et~al.}
\newblock \bibinfo{title}{A neural basis for general intelligence}.
\newblock \emph{\bibinfo{journal}{Science}} \textbf{\bibinfo{volume}{289}},
  \bibinfo{pages}{457--460} (\bibinfo{year}{2000}).

\bibitem{beasley2009rank}
\bibinfo{author}{Beasley, T.~M.}, \bibinfo{author}{Erickson, S.} \&
  \bibinfo{author}{Allison, D.~B.}
\newblock \bibinfo{title}{Rank-based inverse normal transformations are
  increasingly used, but are they merited?}
\newblock \emph{\bibinfo{journal}{Behavior genetics}}
  \textbf{\bibinfo{volume}{39}}, \bibinfo{pages}{580--595}
  (\bibinfo{year}{2009}).

\bibitem{pedregosa2011scikit}
\bibinfo{author}{Pedregosa, F.} \emph{et~al.}
\newblock \bibinfo{title}{Scikit-learn: Machine learning in python}.
\newblock \emph{\bibinfo{journal}{the Journal of machine Learning research}}
  \textbf{\bibinfo{volume}{12}}, \bibinfo{pages}{2825--2830}
  (\bibinfo{year}{2011}).

\bibitem{fortin2018harmonization}
\bibinfo{author}{Fortin, J.-P.} \emph{et~al.}
\newblock \bibinfo{title}{Harmonization of cortical thickness measurements
  across scanners and sites}.
\newblock \emph{\bibinfo{journal}{Neuroimage}} \textbf{\bibinfo{volume}{167}},
  \bibinfo{pages}{104--120} (\bibinfo{year}{2018}).

\bibitem{hoerl1970ridge}
\bibinfo{author}{Hoerl, A.~E.} \& \bibinfo{author}{Kennard, R.~W.}
\newblock \bibinfo{title}{Ridge regression: applications to nonorthogonal
  problems}.
\newblock \emph{\bibinfo{journal}{Technometrics}}
  \textbf{\bibinfo{volume}{12}}, \bibinfo{pages}{69--82}
  (\bibinfo{year}{1970}).

\bibitem{dadi2019benchmarking}
\bibinfo{author}{Dadi, K.} \emph{et~al.}
\newblock \bibinfo{title}{Benchmarking functional connectome-based predictive
  models for resting-state fmri}.
\newblock \emph{\bibinfo{journal}{NeuroImage}} \textbf{\bibinfo{volume}{192}},
  \bibinfo{pages}{115--134} (\bibinfo{year}{2019}).

\bibitem{craddock2013neuro}
\bibinfo{author}{Craddock, C.} \emph{et~al.}
\newblock \bibinfo{title}{The neuro bureau preprocessing initiative: open
  sharing of preprocessed neuroimaging data and derivatives}.
\newblock \emph{\bibinfo{journal}{Frontiers in Neuroinformatics}}
  \textbf{\bibinfo{volume}{7}} (\bibinfo{year}{2013}).

\bibitem{bellec2010multi}
\bibinfo{author}{Bellec, P.}, \bibinfo{author}{Rosa-Neto, P.},
  \bibinfo{author}{Lyttelton, O.~C.}, \bibinfo{author}{Benali, H.} \&
  \bibinfo{author}{Evans, A.~C.}
\newblock \bibinfo{title}{Multi-level bootstrap analysis of stable clusters in
  resting-state fmri}.
\newblock \emph{\bibinfo{journal}{Neuroimage}} \textbf{\bibinfo{volume}{51}},
  \bibinfo{pages}{1126--1139} (\bibinfo{year}{2010}).

\bibitem{huntenburg2017loading}
\bibinfo{author}{Huntenburg, J.} \emph{et~al.}
\newblock \bibinfo{title}{Loading and plotting of cortical surface
  representations in nilearn}.
\newblock \emph{\bibinfo{journal}{Research Ideas and Outcomes}}
  \textbf{\bibinfo{volume}{3}}, \bibinfo{pages}{e12342} (\bibinfo{year}{2017}).

\bibitem{esteve2015big}
\bibinfo{author}{Est{\`e}ve, L.}
\newblock \bibinfo{title}{Big data in practice: the example of nilearn for
  mining brain imaging data}.
\newblock In \emph{\bibinfo{booktitle}{Scipy 2015}} (\bibinfo{year}{2015}).

\bibitem{power2014methods}
\bibinfo{author}{Power, J.~D.} \emph{et~al.}
\newblock \bibinfo{title}{Methods to detect, characterize, and remove motion
  artifact in resting state fmri}.
\newblock \emph{\bibinfo{journal}{Neuroimage}} \textbf{\bibinfo{volume}{84}},
  \bibinfo{pages}{320--341} (\bibinfo{year}{2014}).

\end{thebibliography}

\newpage
\section{Supplementary Material}
\beginsupplement

\hypertarget{an-example-of-model-non-linearity-in-the-presence-of-a-linear-predictor.}{%
\section{An example of model non-linearity in the presence of a linear
predictor.}\label{an-example-of-model-non-linearity-in-the-presence-of-a-linear-predictor.}}
\label{sup:nomlinviol}

Notebook available on-line:

https://github.com/spisakt/mlconfound\_manuscript/tree/main/simulated/normality\_and\_linearity\_violation.ipynb

\begin{lstlisting}[language=Python]
import numpy as np
import seaborn as sns
import matplotlib.pyplot as plt
from sklearn.linear_model import Ridge
sns.set(rc={"figure.figsize":(6, 2)})
sns.set_style("white")
\end{lstlisting}

\hypertarget{five-features-one-linear-four-sigmoid}{%
\subsubsection{Five features: one linear, four
sigmoid}\label{five-features-one-linear-four-sigmoid}}

\begin{lstlisting}[language=Python]
n=300
p=5
rng = np.random.default_rng(42)
y = np.arange(n)
y = (y - y.mean())/y.std()

X_true = np.repeat(y, p).reshape(n,p)
for i in range(1,4):
    X_true[:,i] = np.tanh(X_true[:,i]*2)

X=X_true + rng.normal(0,0.1, (n,p))

for i in range(4):
    sns.lineplot(x=y, y=X_true[:,i])
    sns.scatterplot(x=y, y=X[:,i])
    plt.show()
\end{lstlisting}

\begin{figure}[H]
\centering
\includegraphics[width=0.15\paperwidth]{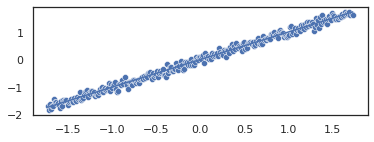}
\includegraphics[width=0.15\paperwidth]{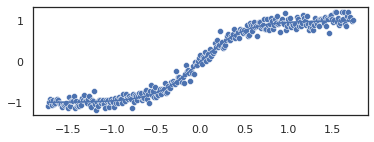}
\includegraphics[width=0.15\paperwidth]{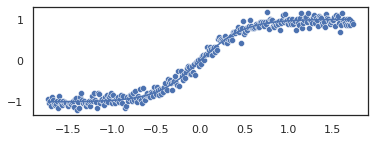}
\includegraphics[width=0.15\paperwidth]{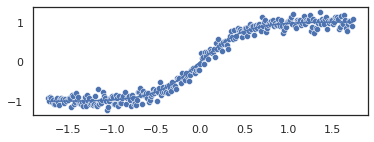}
\end{figure}

\hypertarget{model-predictions-without-regularization}{%
\subsubsection{\texorpdfstring{Model predictions \textbf{without}
regularization}{Model predictions without regularization}}\label{model-predictions-without-regularization}}

\begin{lstlisting}[language=Python]
model = Ridge(alpha=0)
model.fit(y=y, X=X)
yhat = model.predict(X)
sns.scatterplot(x=y, y=yhat, alpha=0.3)
model.coef_
\end{lstlisting}

\begin{lstlisting}
array([ 0.47902429,  0.00946499,  0.07378478, -0.02828672,  0.47070535])
\end{lstlisting}

\begin{figure}[H]
\centering
\includegraphics[width=0.5\paperwidth]{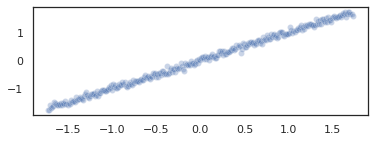}
\end{figure}

\hypertarget{model-predictions-with-regularization}{%
\subsubsection{\texorpdfstring{Model predictions \textbf{with}
regularization}{Model predictions with regularization}}\label{model-predictions-with-regularization}}

\begin{lstlisting}[language=Python]
model = Ridge(alpha=2000)
model.fit(y=y, X=X)
yhat = model.predict(X)
sns.scatterplot(x=y, y=yhat, alpha=0.3)
model.coef_

\end{lstlisting}

\begin{lstlisting}
array([0.09472368, 0.07472972, 0.07338482, 0.07440682, 0.09463765])
\end{lstlisting}

\begin{figure}[H]
\centering
\includegraphics[width=0.5\paperwidth]{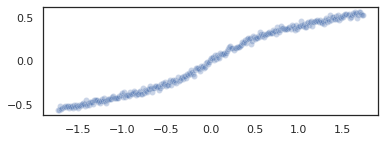}
\end{figure}

\section{Full confounder test}
\label{sup:full-test}

The full confoudner test generates a null-distribution for an arbitrary predefined test statistic $T(\y,\yhat,\c)$ by sampling permutation based "copies" of $\y$,

$$\y_i^{(j)} \sim Q(\cdot|c_i) \ $$

where, $Q(.|c)$ denotes the conditional distribution of $\y$ given $\c=c$ and $j=1,\dots, m$ indexes the "copy" of $\y$ so that $\y^{(i)} = (y_1^{(j)}, \dots, y_n^{(j)})$ is a permutation of the original vector $\y = (y_1, \dots, y_n)$. This mechanism creates copies $\y^{(1)}, \dots ,\y^{(m)}$ that are exchangeable with the original vector $\y$ under the null hypothesis that $\yhat \independent \y | \c$.

Under the null hypothesis, the triples $$(\y,\yhat,\c), (\y^{(1)}, \yhat, \c),\dots, (\y^{(1)}, \yhat, \c)$$ are all identically distributed and exchangeable, and so are the 
$$T(\y,\yhat,\c), T(\y^{(1)}, \yhat, \c),\dots,T(\y^{(m)}, \yhat, \c)$$
test statistics, as well.

The p-value under the null hypothesis is then obtained as
$$ p= \frac{\sum_{j=1}^m \mathbb{1} \{T(\y^{(j)}, \yhat, \c) \geq T(\y, \yhat, \c) \}  }{m}$$

Let $S_n$ denote the set of all permutations on the indices $\{1,\dots,n\}$ and $\y_\pi = (y_{\pi_1}, \dots, y_{\pi_n})$ the vector $\y$ with its elements reordered according to the permutation $\pi \in S_i$.
The permutation-based copies of $\y$ are then of $\y^{(j)} = \y_{\pi^{(j)}}$ which are drawn so that:

    $$\mathbb{P}(\pi^{(j)} = \pi | \y,\yhat,\c) = \frac{q^n(\y_\pi | \c)}{\sum_{\pi' \in S_n} q^n(\y_{\pi'} | \c)}$$

that is, according to the $q^n(\cdot|\c) := q(\cdot | c_1) \dots q(\cdot|c_n)$ product density corresponding to the conditional distribution $Q(\cdot|\c)$. Note that Eq. \ref{eq-pperm} does not necessarily assume a continuous distribution.
For a verification of the valid type I error control of this approach refer to Theorem 1 in \citep{berrett2020conditional}.

\begin{figure}[H]
  \centering
  \fbox{
   \includegraphics[width=0.36\paperwidth]{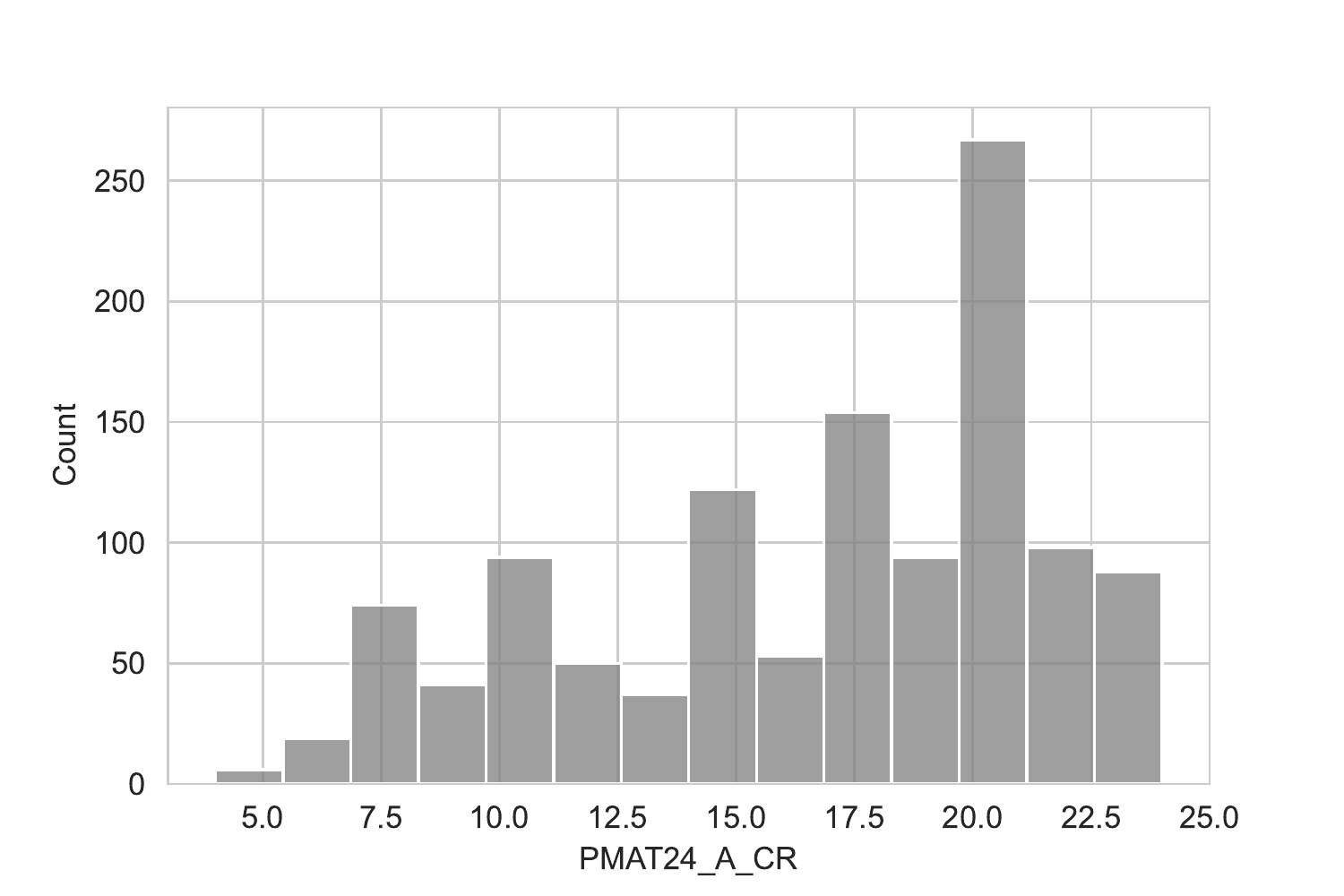}
   \includegraphics[width=0.36\paperwidth]{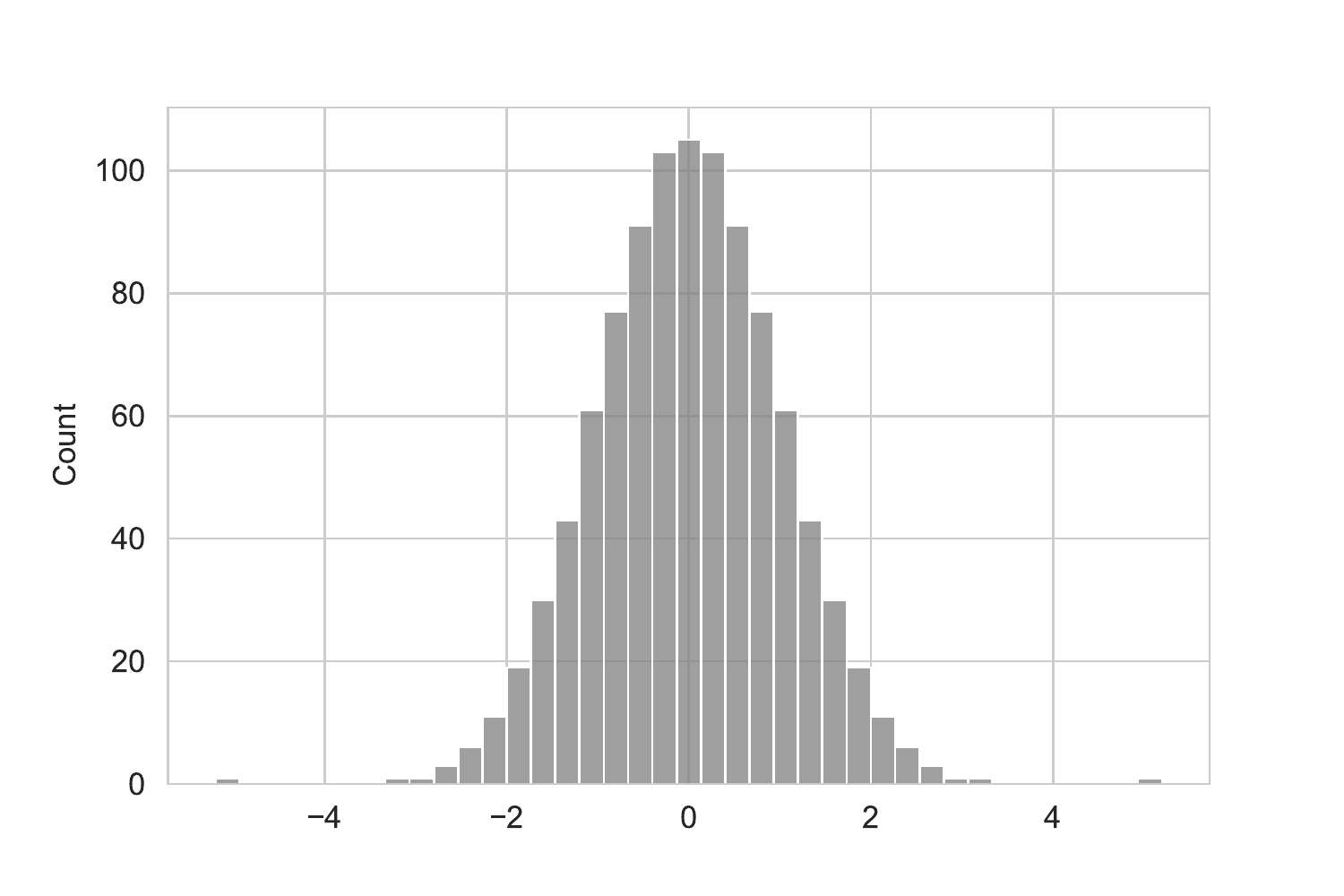}
   }
  \caption{Histogram of fluid intelligence score in the HPC dataset, before (left) and after (right) quantile transformation.}
  \label{fig:hcp-hist}
\end{figure}

\begin{figure}[H]
  \centering
  \fbox{
   \includegraphics[width=0.36\paperwidth]{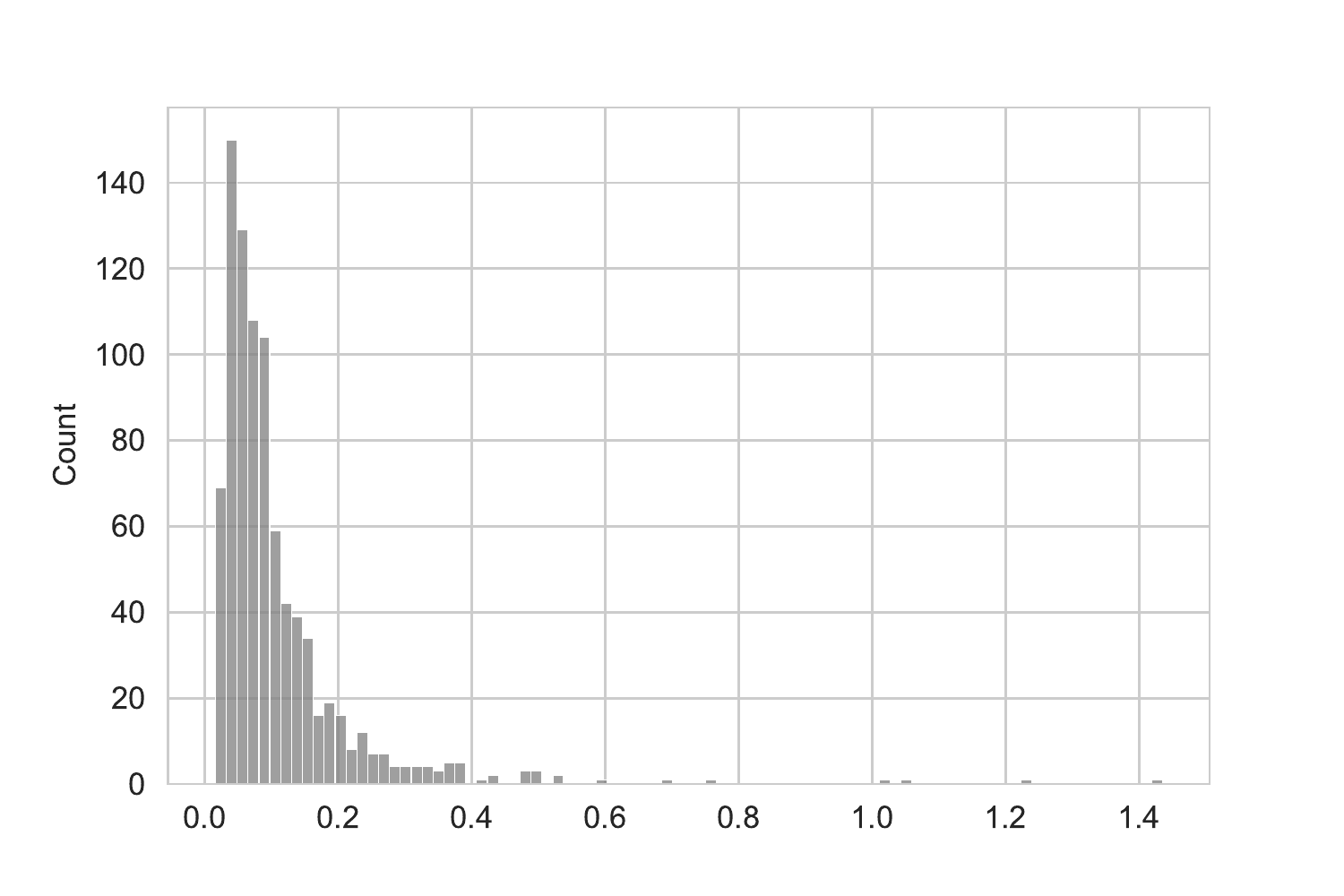}
   \includegraphics[width=0.36\paperwidth]{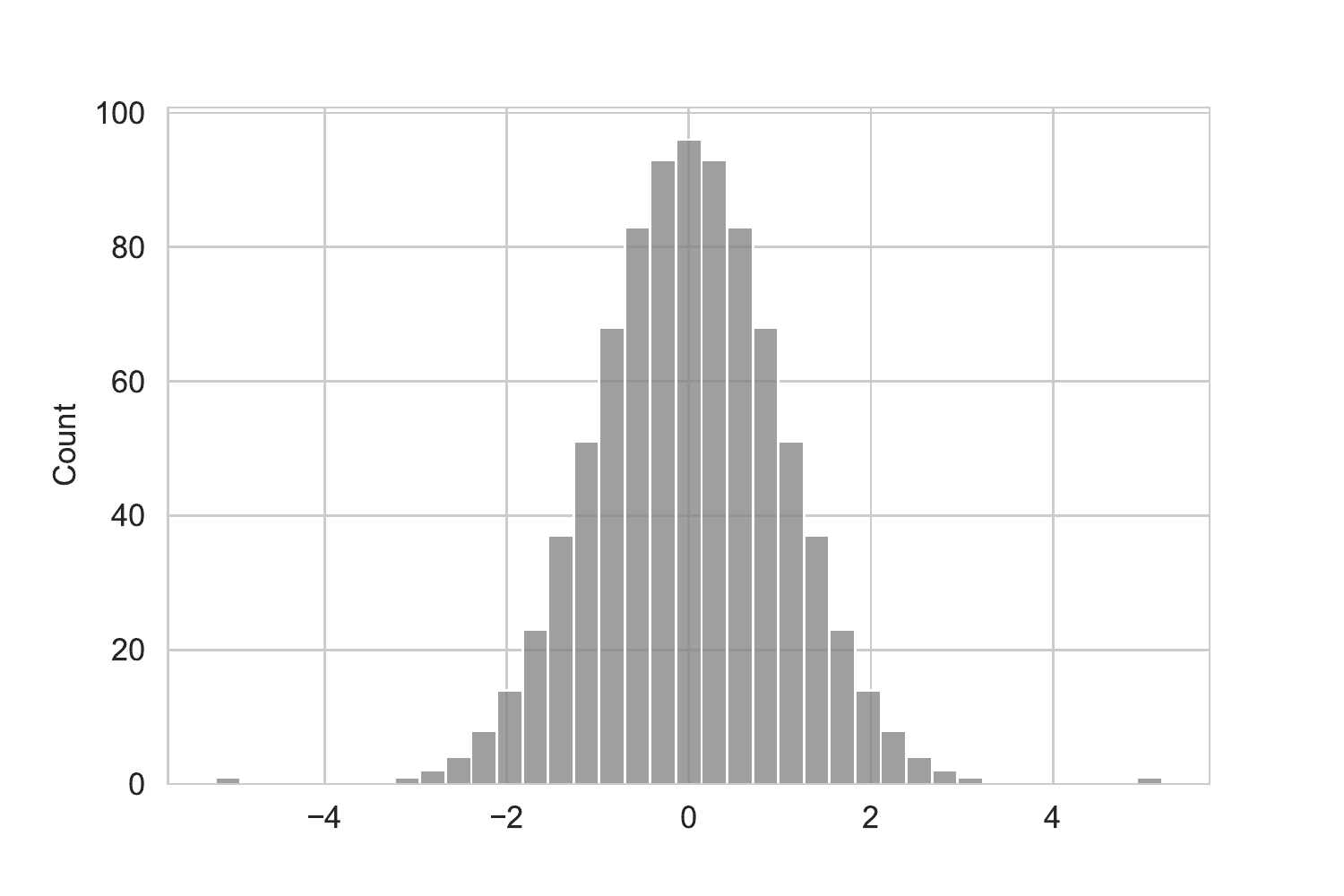}
   }
  \caption{Histogram of mean framewise displacement in the ABIDE dataset, before (left) and after (right) quantile transformation.}
  \label{fig:abide-hist}
\end{figure}

\end{document}